%% file: main.tex
\documentclass{./styles/svproc}
\usepackage[a4paper]{geometry}
\usepackage{geometry}
\usepackage{wrapfig}
\usepackage[font=small,labelfont=bf]{caption} 
\usepackage{graphicx}
\usepackage{amssymb}
\usepackage{amsmath}
\usepackage{url}

\usepackage[colorlinks]{hyperref}
\usepackage{subcaption}

\usepackage{algorithm}
\usepackage{algorithmic}
\usepackage[dvipsnames]{xcolor}
\usepackage{cite}
\usepackage{placeins}

\begin{document}
\mainmatter              % start of a contribution
%
% \title{MERGE :Manipulation and Exploration with Reinforced Guidance and Execution }
% \title{NAVINACT: Merging Navigation and Interaction Networks for Efficient Robotic Learning}
% \title{NAVINACT: Combining Navigation and Imitation Learning for Bootstrapping Reinforcement Learning}
\title{PLANRL: A Motion Planning and Imitation Learning Framework to Bootstrap Reinforcement Learning}
\titlerunning{PLANRL}  % abbreviated title (for running head)
%                                   \toctitle is used
\newcommand{\NAVINACT}{\textsc{PLANRL}}
\newcommand{\todo}[1]{\textcolor{red}{TODO:#1}}
\bibliographystyle{./styles/bibtex/spbasic} % or another style suitable for Springer
% \bibliography{bibliography} % This tells LaTeX the .bib file location
\author{Amisha Bhaskar \and Zahiruddin Mahammad \and
Sachin R Jadhav  \and \\ 
Pratap Tokekar}
\authorrunning{Bhaskar et al.,} % abbreviated author list (for running head)
\tocauthor{Amisha Bhaskar, Zahir Mahammed, Sachin Jadhav, and Pratap Tokekar}
\institute{University of Maryland, College Park MD 20742, USA\\
\email{\{amishab,zahirmd,sjd3333,tokekar\}@umd.edu}}
% ,\\ WWW home page:
% \texttt{http://users/\homedir iekeland/web/welcome.html}
% \and
% Universit\'{e} de Paris-Sud,
% Laboratoire d'Analyse Num\'{e}rique, B\^{a}timent 425,\\
% F-91405 Orsay Cedex, France}

\maketitle              % typeset the title of the contribution
\thispagestyle{empty}
\pagestyle{empty}
%%%%%%%%%%%%%%%%%%%%%%%%%%%%%%%%%%%%%%%%%%%%%%%%%%%%%%%%%%%%%%%%%%%%%%%%%%%%%%%%

\input{root/sections/0.abstract}

\input{root/sections/1.introduction}

\input{root/sections/2.related_work}
% \input{sections/3.preliminaries}
\input{root/sections/4.proposed_approach}

\input{root/sections/5.experiments}

\input{root/sections/6.conclusion}

%%%%%%%%%%%%%%%%%%%%%%%%%%%%%%%%%%%%%%%%%%%%%%%%%%%%%%%%%%%%%%%%%%%%%%%%%%%%%%%%


\end{document}

%% file: root/sections/0.abstract.tex
\begin{abstract}
Reinforcement Learning (RL) has shown remarkable progress in simulation environments, yet its application to real-world robotic tasks remains limited due to challenges in exploration and generalization. To address these issues, we introduce \NAVINACT{}, a framework that chooses when the robot should use classical motion planning and when it should learn a policy. To further improve the efficiency in exploration, we use imitation data to bootstrap the exploration. \NAVINACT{} dynamically switches between 
two modes of operation: reaching a waypoint using classical techniques when away from the objects and reinforcement learning for fine-grained manipulation control when about to interact with objects. \NAVINACT{}  architecture is composed of ModeNet for mode classification, NavNet for waypoint prediction, and InteractNet for precise manipulation. By combining the strengths of RL and Imitation Learning (IL), \NAVINACT{} improves sample efficiency and mitigates distribution shift, ensuring robust task execution. We evaluate our approach across multiple challenging simulation environments and real-world tasks, demonstrating superior performance in terms of adaptability, efficiency, and generalization compared to existing methods.  In simulations, \NAVINACT{} surpasses baseline methods by 10-15\% in training success rates at 30k samples and by 30-40\% during evaluation phases. In real-world scenarios, it demonstrates a 30-40\% higher success rate on simpler tasks compared to baselines and uniquely succeeds in complex, two-stage manipulation tasks.
 Datasets and supplementary materials can be found on our \href{https://raaslab.org/projects/NAVINACT/}{website}.
% This work advances the state of robotic learning by enabling more autonomous, efficient, and adaptable robotic systems.
\keywords{Reinforcement Learning, Imitation Learning, Efficiency, Generalization, Manipulation}
\end{abstract}

%% file: root/sections/1.introduction.tex
\section{Introduction}
\label{section:introduction}
\begin{figure}[h]
    \centering
    \includegraphics[width=\linewidth]{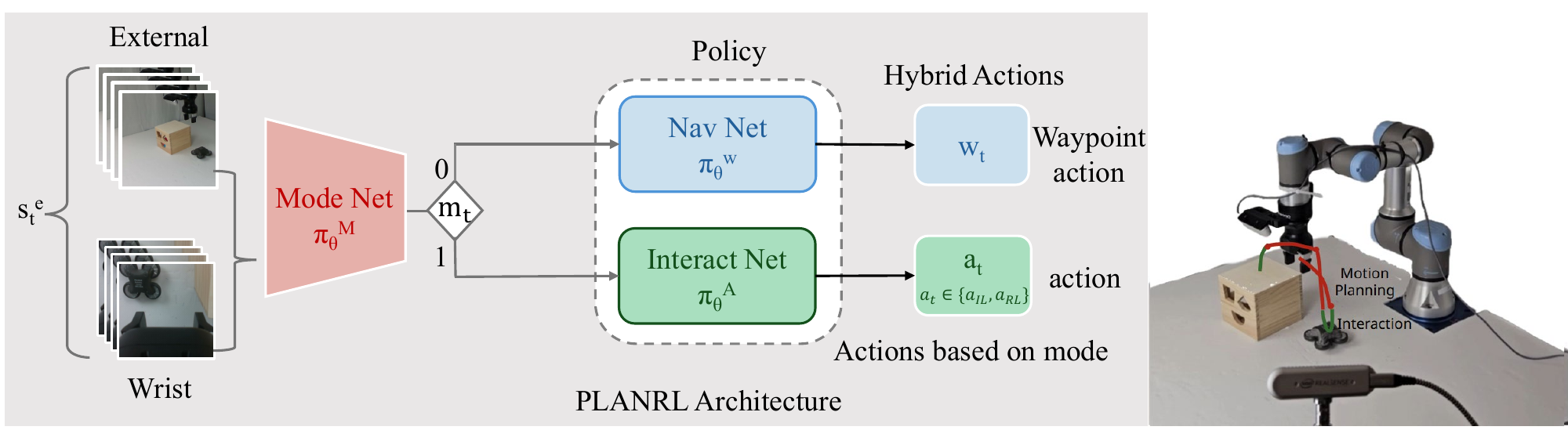}
    \caption{Architecture of \NAVINACT{}: During training, \NAVINACT{} learns to predict waypoints, low-level actions, and the operational mode at each time step. One network (InteractNet) predicts the low-level action \( a_t \) and the other network (ModeNet) predicts mode \( m_t \). A separate network (NavNet) predicts the high-level waypoint \( w_t \). At test time, the system samples \( m_t \) and either moves to a waypoint (when \( m_t = 0 \)) using the predicted waypoint or follows a dense action (when \( m_t = 1 \)). The architecture allows for dynamic switching between motion-planning and interaction modes, facilitating robust performance in complex tasks. An example of how motion planning and interaction modes are integrated during execution is shown on the right.
}
    \label{architecture}
\end{figure}

In recent years, reinforcement learning (RL) has made significant strides, achieving remarkable success across various domains \cite{silver,oriol,fair}. 
Despite advances in RL, learning behaviors efficiently that generalize to new settings is still a challenge.
RL methods are often constrained to short-horizon tasks due to the inherent challenges of exploration in high-dimensional, continuous action spaces, which are characteristic of robotic applications. In these scenarios, reward signals tend to be sparse, making it difficult for a randomly initialized policy to receive any learning signals, especially when the task's completion is far from the initial state. Additionally, RL typically requires a large number of samples to converge, a requirement that is impractical in real-world environments where large-scale parallel simulation is not feasible.  The exploration challenges are further exacerbated when tasks demand both high-level strategic planning and precise low-level control, making it difficult for a single RL model to effectively learn and execute the nuanced decompositions needed for complex manipulation tasks.

To enhance sample efficiency and generalizability, researchers have explored imitation learning (IL) as a means to accelerate learning by leveraging demonstration data.  In hierarchical frameworks like LAVA \cite{lava}, IL is used to parameterize low-level policies based on high-level visual inputs, enabling the system to effectively handle varying types and positions of food during robotic-assisted feeding tasks. However, IL alone can lead to distributional shift issues when applied outside the training context. Our focus in this paper is to improve the sample complexity and generalizability of robot learning specifically for manipulation tasks. Our key idea is to improve the exploration process relying on classical motion planning algorithms and imitation data. Specifically, we use motion planning when the robot is ``away" from objects and apply RL only when the robot is ``near" objects during interactions that require fine control. This strategy reduces the complexity of the learning problem and enhances sample efficiency. Further improvements in sample efficiency can be achieved through imitation learning (IL), though IL alone can raise generalizability issues, which we address by bootstrapping IL with RL.

Our novel approach leverages the hierarchical nature of robotic tasks by implementing a dynamic policy framework that switches between moving to high-level waypoints and engaging in fine-grained manipulation. We build on HYDRA \cite{hydra} which introduced the idea of switching between sparse (classical motion planning) and dense (learning) modes. However, the dense mode in HYDRA uses IL which is prone to distributional shift near objects, our method integrates IL with RL, mitigating these shifts and improving task robustness.

While RL is effective in optimizing policies through exploration, it struggles with sparse and delayed rewards. Conversely, IL can expedite learning by providing optimal or near-optimal actions, but it is constrained by static training datasets and the risk of distributional shift when applied to novel contexts. To address these limitations, we combine the strengths of both RL and IL. Methods such as Imitation Bootstrapped Reinforcement Learning (IBRL) \cite{ibrl}, \cite{pegMARL} and Reinforcement Learning from Prior Data (RLPD) \cite{rlpd} demonstrate this integration, achieving better results compared to using IL or RL alone. However, these methods still require many interaction steps to converge. By incorporating IBRL with mode prediction through ModeNet and waypoint prediction through NavNet, our approach achieves results much faster, significantly improving sample efficiency and performance in complex robotic tasks.

We introduce \NAVINACT{} (\textbf{NAV}igation and \textbf{IN}ter\textbf{ACT}ion), a framework that integrates motion-planning and imitation learning to bootstrap reinforcement learning. Our approach comprises three key components: NavNet for strategic waypoint planning based on visual inputs; ModeNet that dynamically switches between motion-planning and interaction modes; and InteractNet that performs fine-grained manipulation tasks with precision as shown in Fig. \ref{modes}. The key contributions of this paper are:
\begin{itemize}
    \item We present \NAVINACT{}, a hierarchical policy framework designed to integrate motion-planning and interaction for complex robotic manipulation tasks.
    \item Our method demonstrates adaptability and robustness across various challenging environments, effectively handling both high-level planning and detailed execution.
    \item We introduce a generalized mode prediction and waypoint prediction network, reducing the need for task-specific mode labeling and enhancing the system's scalability.
    \item We thoroughly tested \NAVINACT{} in both simulated and real-world settings using a UR3e robot equipped with a Robotiq Hand-e gripper and Realsense cameras. The framework achieved impressive success rates: 85±5\% during simulation training and 80±5\% in evaluations with unseen environments. In real-world tests, \NAVINACT{} achieved up to 90\% success on simpler tasks and 75\% on complex two-stage tasks within 30k samples, highlighting its efficiency and adaptability.
\end{itemize}

%% file: root/sections/2.related_work.tex
\section{Related Work}
\label{section:related_research}

This section explores substantial contributions in reinforcement learning (RL) aimed at enhancing sample efficiency and generalizability, with a focus on methods integrating human demonstrations and strategies to counter distribution shifts in imitation learning (IL). We also examine how leveraging reference policies can optimize RL outcomes.

\subsection{Incorporating Model Priors and Refining Action Spaces}
Research has increasingly concentrated on embedding structural priors within models to stabilize performance across variable conditions. Object-centric representations and pretrained state models from diverse task datasets have proven to enhance generalization and efficiency \cite{zhu, s_karamcheti, s_nair}. Approaches like \cite{hydra} HYDRA have refined action spaces to align with these insights. However, \NAVINACT{} goes a step further by dynamically switching action modes, integrating both high-level strategic planning and precise low-level task executions. This not only enhances motion-planning and manipulation capabilities in intricate settings but also addresses the challenges of generalization and extensive data requirements found in temporal action abstractions and parameterized motion primitives \cite{pertsch, lynch, belkhale, shridhar, akgun, chi}. Building on the foundations laid by previous models, \NAVINACT{} effectively minimizes errors and advances the capabilities of action representation in complex environments, making it a significant advancement in both model integration and action space refinement.

\subsection{Planning-Driven Reinforcement Learning}
Integrating motion planning with RL has been explored to combine the benefits of both paradigms \cite{lee_et_al,yamada,cheng,xia_et_at,james_davison,james_et_al,liu_et_al}. For example, GUAPO \cite{lee_et_al} focuses on single-stage tasks, maintaining RL agents within low pose-estimator uncertainty areas, while Plan-Seq-Learn \cite{planseq} decomposes complex tasks into sequences of subtasks guided by large language models (LLMs). However, these methods often come with high inference costs or inefficiencies in task decomposition. Our \NAVINACT{} method contrasts these approaches by offering an end-to-end learning solution that is not only faster but also more adaptable to real-time applications, with the potential for further enhancement through LLM integration.

\subsection{Reinforcement Learning with Prior Demonstrations} Sample-efficient RL often struggles in sparse reward environments, where initial learning cues are vital. Integrating prior data or human demonstrations into the replay buffer—an approach employed in methods like Reinforcement Learning from Prior Data (RLPD)\cite{rlpd} and IBRL \cite{ibrl}—has proven effective. These methods enhance RL algorithms by oversampling demonstrations during training, significantly improving performance in continuous control domains with techniques like normalization and Q-ensembling [2, 28]. Our \NAVINACT{} framework builds on these principles by dynamically adjusting its strategy based on environmental cues and task demands, using an architecture that optimizes both motion-planning and interaction. This approach not only leverages prior demonstrations for improved initial guidance but also adapts in real-time to enhance task execution and generalization in complex robotic manipulation tasks.

\subsection{Advancing RL Sample Efficiency}
Recent advancements in RL have prioritized enhancing sample efficiency through innovative regularization strategies. Techniques like RED-Q \cite{red_q} and DropoutQ \cite{dropout_q} apply regularization to the Q-function, utilizing ensembling or dropout to achieve higher update-to-data (UTD) ratios and expedited convergence. These approaches prove particularly effective in state-based RL, while in pixel-based settings, strategies such as image augmentation with random shifts \cite{denis} boost performance without additional computational demands. Building upon these concepts, the IBRL \cite{ibrl} framework combines imitation learning (IL) with reinforcement learning (RL) by leveraging a pre-trained IL policy to guide RL training from the start, allowing the RL agent to benefit from high-quality actions provided by the IL policy. This integration significantly accelerates both exploration and training efficiency. However, IBRL still necessitates frequent updates to achieve optimal performance levels. In contrast, our \NAVINACT{} method refines this approach by significantly reducing the number of updates required, effectively merging the benefits of IL, RL as well as mode switching and motion-planning to heighten sample efficiency and streamline learning processes.
% Recent innovations have focused on boosting RL's sample efficiency through techniques like RED-Q and DropoutQ, which employ regularization strategies to expedite learning \cite{red_q, dropout_q}. While effective, these approaches are generally confined to state-based scenarios. NAVINACT further refines this by efficiently merging IL insights with RL tactics, significantly reducing the need for extensive iterative updates and improving learning outcomes with fewer data interactions.

% These explorations underscore NAVINACT’s pivotal role in pushing the boundaries of what's achievable in adaptive robotic tasks, setting new benchmarks for both efficiency and adaptability in robotic systems.

%% file: root/sections/4.proposed_approach.tex
\section{\NAVINACT{}}
\label{section:proposed_approach}

We consider a standard Markov decision process (MDP) consisting of state space $s \in \mathcal{S}$, continuous action space $\mathcal{A} = [-1, 1]^d$, deterministic state transition function $T : \mathcal{S} \times \mathcal{A} \to \mathcal{S}$, sparse reward function $R : \mathcal{S} \times \mathcal{A} \to \{0, 1\}$ that returns 1 when the task is completed and 0 otherwise, and discount factor $\gamma$.

\begin{figure}[ht]
    \centering
    \begin{minipage}{0.49\textwidth}
        \centering
        \includegraphics[width=\linewidth]{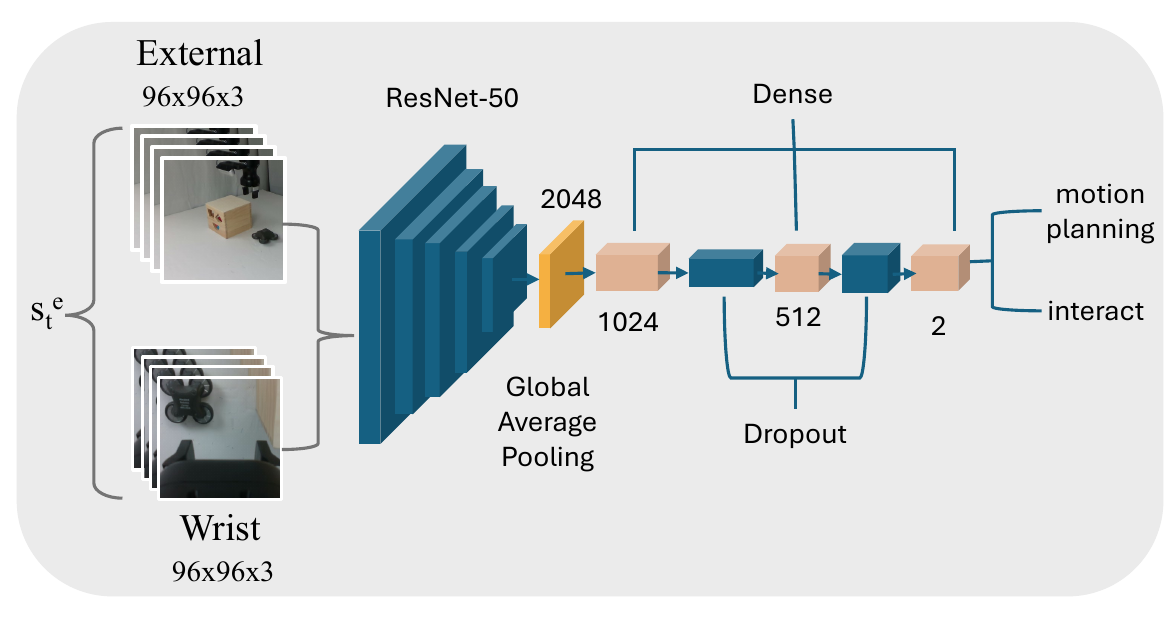}
        \caption{ModeNet architecture designed to classify modes (motion-planning vs. interact) based on visual inputs.}
        \label{modenet}
    \end{minipage}\hfill
    \begin{minipage}{0.49\textwidth}
        \centering
        \includegraphics[width=\linewidth]{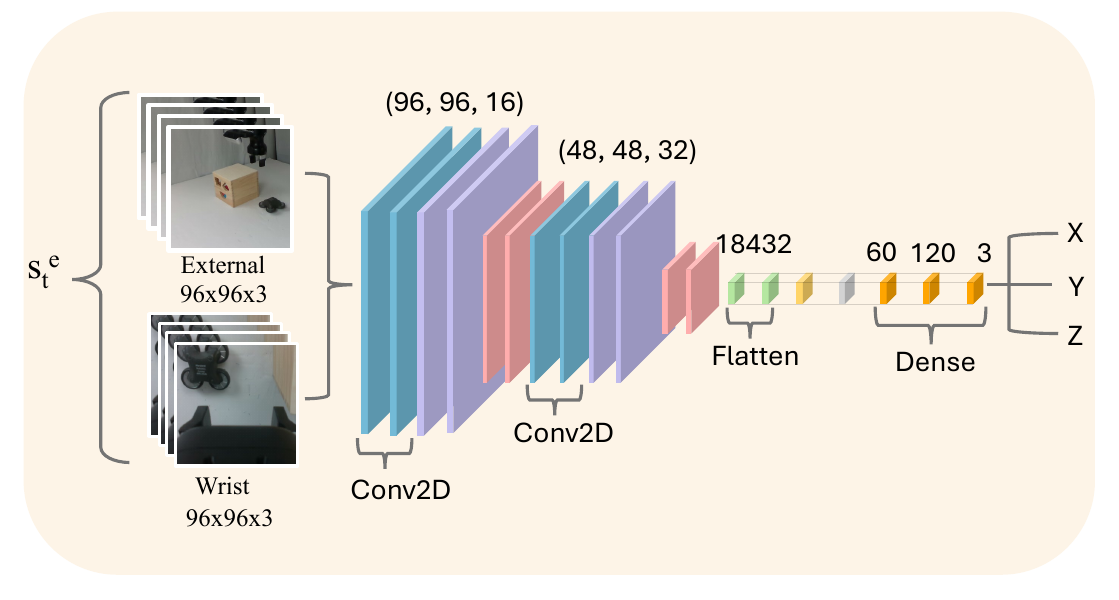}
        \caption{NavNet architecture for predicting
waypoints to guide high-level motion-planning
tasks.}
        \label{navnet}
    \end{minipage}
\end{figure}

% \begin{wrapfigure}{r}{0.5\textwidth}
%     \centering
%     \includegraphics[width=\linewidth]{isrr2024_format/templates/figures/mode_network.pdf}
%     \caption{ModeNet architecture for classifying operational modes (navigate vs. interact) based on visual and proprioceptive inputs.}
%     \label{modenet}
% \end{wrapfigure}
Our proposed method, \NAVINACT{}, features an architecture designed for efficient robotic manipulation. The system is composed of three main components: ModeNet, NavNet, and InteractNet, each contributing to different levels of task execution. The overall architecture, as illustrated in Fig. \ref{architecture}, includes heads $\pi^M_\theta: \mathcal{S} \rightarrow \{0, 1\}$, $\pi^W_\theta: \mathcal{S} \rightarrow W$, and $\pi^A_\theta: \mathcal{S} \rightarrow \mathcal{A}$, corresponding to mode classification, high-level waypoint prediction, and low-level action prediction respectively, where $W$ is robot end-effector position. This design enables dynamic switching between coarse and fine-grained control during task execution.

\subsection{ModeNet: A Vision-Based Mode Classification Network}
ModeNet is a deep learning architecture that classifies the operational mode of the robot based on visual input. It determines whether the robot should operate in motion-planning mode, or in interaction mode, as shown in Fig. \ref{modes}.\\
% \textbf{Network Architecture:}
ModeNet utilizes a ResNet backbone as shown in Fig. \ref{modenet}. The architecture consists of a feature extractor followed by a mode classifier. The feature extractor employs a pre-trained ResNet model to capture high-level spatial and contextual features from input images, while the mode classifier processes these features to predict the robot's operational mode.

\subsection{NavNet: A Vision-Based Waypoint Prediction Network}

NavNet is designed to predict waypoints for robotic manipulation tasks by processing visual inputs. It generates high-level action outputs that guides the robot towards the target objects in the environment. NavNet uses a convolutional neural network (CNN) for visual processing, effectively handling the image data to produce actionable waypoints, Fig. \ref{navnet}. Once we have the predicted waypoint, we use a sampling-based motion planner, AIT* \cite{AT}, to perform motion planning as it requires minimal setup—just collision checking—and performs well in planning tasks.
\begin{algorithm}[t]
\caption{\NAVINACT{}. Major modifications with framework highlighted in \textcolor{RoyalBlue}{blue}.}
\label{alg:navinact}
\begin{algorithmic}[1]
\STATE \textbf{Hyperparameters:} Number of critics $E$, number of critic updates $G$, update frequency $U$, exploration std $\sigma$, noise clip $c$, Complexity Function \( g \)
\STATE Train imitation policy $\mu_\psi$ on demonstrations $D = \{\xi_1, \dots, \xi_n\}$ using the selected IL algorithm.
\color{RoyalBlue}\STATE \textcolor{RoyalBlue}{Train ModeNet on dataset $D_{\text{mode}}$ to classify mode (motion-planning vs interaction).}
\color{RoyalBlue}\STATE \textcolor{RoyalBlue}{Train NavNet on dataset $D_{\text{nav}}$ to predict waypoints for motion-planning tasks.}
\color{Black}\STATE Initialize policy $\pi_\theta$, target policy $\pi_{\theta'}$, and critics $Q_\phi$, target critics $Q_{\phi'}$ for $i = 1, \dots, E$
\STATE Initialize replay buffer $B$ with demonstrations $\{\xi_1, \dots, \xi_n\}$
\FOR{$t = 1$ \TO $N$}  % Changed num_rl_steps to N
    \STATE Observe current state $s_t$ from the environment
    \color{RoyalBlue}\STATE \textcolor{RoyalBlue}{Determine mode \( m_t \) using ModeNet.}
    \IF{mode $m_t$ is 0}
        \STATE Predict waypoint $w_t$ using NavNet
        \STATE Generate motion-planning action $a^{\text{Nav}}_t$ based on waypoint $w_t$
        \STATE Execute motion-planning action $a^{\text{Nav}}_t$
    \ELSE
        \color{Black}\STATE Compute IL action $a^{\text{IL}}_t \sim \mu_\psi(s_t)$ and RL action $a^{\text{RL}}_t = \pi_\theta(s_t) + \epsilon$, where $\epsilon \sim N(0, \sigma^2)$
        \STATE Sample a set $K$ of 2 indices from $\{1, 2, \dots, E\}$
        \STATE Select action $a_t$ with higher Q-value from $\{a^{\text{IL}}, a^{\text{RL}}\}$
        \STATE Execute action $a_t$
        \STATE Store transition $(s_t, a_t, r_t, s_{t+1})$ in replay buffer $B$
        \IF{$t \% U = 0$}  % Changed the condition check
            \STATE Perform TD3 update using minibatches from replay buffer $B$ \cite{td3}
        \ENDIF
    \ENDIF
\ENDFOR
\end{algorithmic}
\end{algorithm}
\FloatBarrier

% \begin{wrapfigure}{}{0.5\textwidth}
%     \centering
%     \includegraphics[width=\linewidth]{isrr2024_format/templates/figures/nav_network.pdf}
%     \caption{NavNet architecture for predicting waypoints.}
%     % \vspace{-20pt}
%     \label{navnet}
% \end{wrapfigure}

\subsection{InteractNet: Low-Level Action Prediction}
InteractNet integrates imitation learning (IL) with reinforcement learning (RL) to enhance sample efficiency in robotic task execution. Initially, it trains using expert demonstrations to establish a baseline IL policy. This IL policy guides the RL training phase by recommending high-quality actions, aiding in effective learning through techniques like Temporal Difference (TD) and off-policy RL methods such as TD3. Actions are selected based on the highest Q-values, comparing inputs from both IL and RL strategies.

% \begin{wrapfigure}{r}{0.5\textwidth}
%     \centering
%     \includegraphics[width=\linewidth]{isrr2024_format/templates/figures/assembly_bc_bar.pdf}
%     \caption{Analysis of action selection from Behavior Cloning (BC) policy during \NAVINACT{} training. The figures show the proportion of actions taken from the BC and RL policy across different tasks, providing insight into the system's reliance on BC during training.}
%     \vspace{-50pt}
%     \label{bc_vs_rl}
% \end{wrapfigure}

\begin{wrapfigure}{r}{0.5\textwidth}
    \centering
    \vspace{-10pt}
    \begin{subfigure}{\linewidth}
        \includegraphics[width=\textwidth]{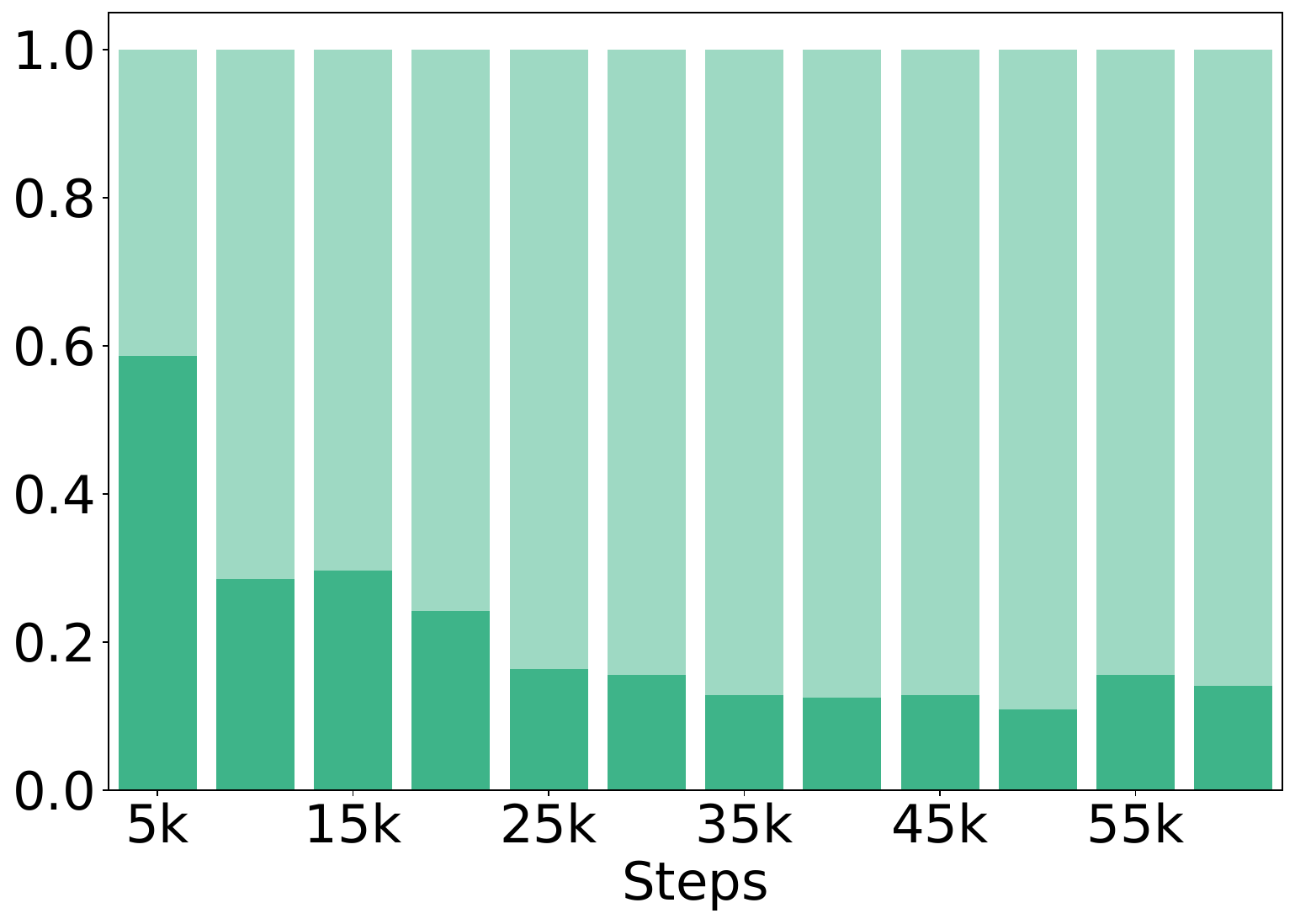}
        % \caption{Main figure caption.}
    \end{subfigure}
    
    \begin{subfigure}{0.7\linewidth}
        \includegraphics[width=\textwidth]{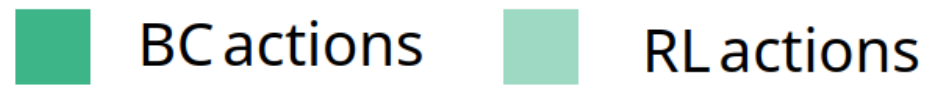} % Update this path with your subfigure's path
        % \caption{Subfigure caption.}
    \end{subfigure}
    \caption{Analysis of action selection from Behavior Cloning (BC) policy during NAVINACT training. The figures show the proportion of actions taken from the BC and RL policy across different tasks, providing insight into the system's reliance on BC during training.}
    \label{bc_vs_rl}
    \vspace{-30pt}
\end{wrapfigure}

Fig. \ref{bc_vs_rl} illustrates how often IL actions are chosen throughout training.
Early on, the predominance of RL actions is noticeable due to initially overestimated Q-values. As the system updates and learns from real performance data, IL actions become more frequent, recognizing their effectiveness from expert demonstrations in the replay buffer. Over time, although reliance on IL actions decreases as the RL policy matures, they remain essential in complex tasks where RL alone is insufficient for full policy convergence. This consistent integration of IL actions at all stages underscores their critical role in ensuring robust and efficient decision-making within \NAVINACT{}.

% \subsection{System Integration and Execution}

% At test time, NAVINACT dynamically switches between mode prediction, waypoint navigation, and low-level action execution. ModeNet determines whether the robot should navigate to a waypoint or perform an interaction. NavNet guides the robot to the target objects, while InteractNet handles precise manipulation tasks, ensuring smooth and efficient task completion.

%% file: root/sections/5.experiments.tex
\section{Experiments}
\label{section:experiments}

% \begin{wrapfigure}{r}{0.5\textwidth}
%     \centering
%     \includegraphics[width=\linewidth]{isrr2024_format/templates/figures/boxclose_gripper_rand.pdf}
%     \caption{This figure displays the evaluation setups for MetaWorld task, with randomized initial gripper positions. The figure illustrates how the gripper's starting position varies in the environments, testing NAVINACT's ability to adapt to different initial conditions.}
%     \vspace{-80pt}
%     \label{gripper_rand_fig}
% \end{wrapfigure} 

 In this section, we present a comprehensive evaluation of \NAVINACT{}, focusing on its sample efficiency, generalizability, and performance in real-world scenarios. We conducted experiments in three challenging simulation environments (Fig. \ref{sim_results}) and two complex real-world tasks (Fig. \ref{real_world_result}).
\subsection{Training and evaluation of ModeNet and NavNet}

% \subsubsection{ModeNet}
\begin{figure}[h]
    \centering
        \includegraphics[width=\linewidth]{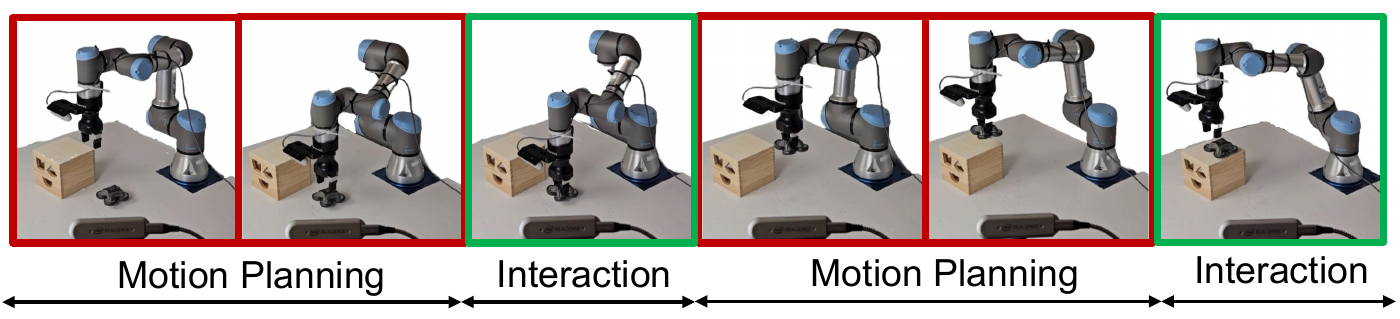}
        \caption{Sequence of images illustrating ModeNet's predicted modes during trajectory execution.}
        \label{modes}
        % \vspace{-10pt}
\end{figure}
\begin{figure}[h]
    \centering
    \begin{subfigure}{0.32\textwidth}
        % \hspace{0.08\textwidth}
        % \raisebox{0.15\height}{\includegraphics[width=0.95\textwidth, height=0.15\textheight]
        \includegraphics[width=\linewidth]
        {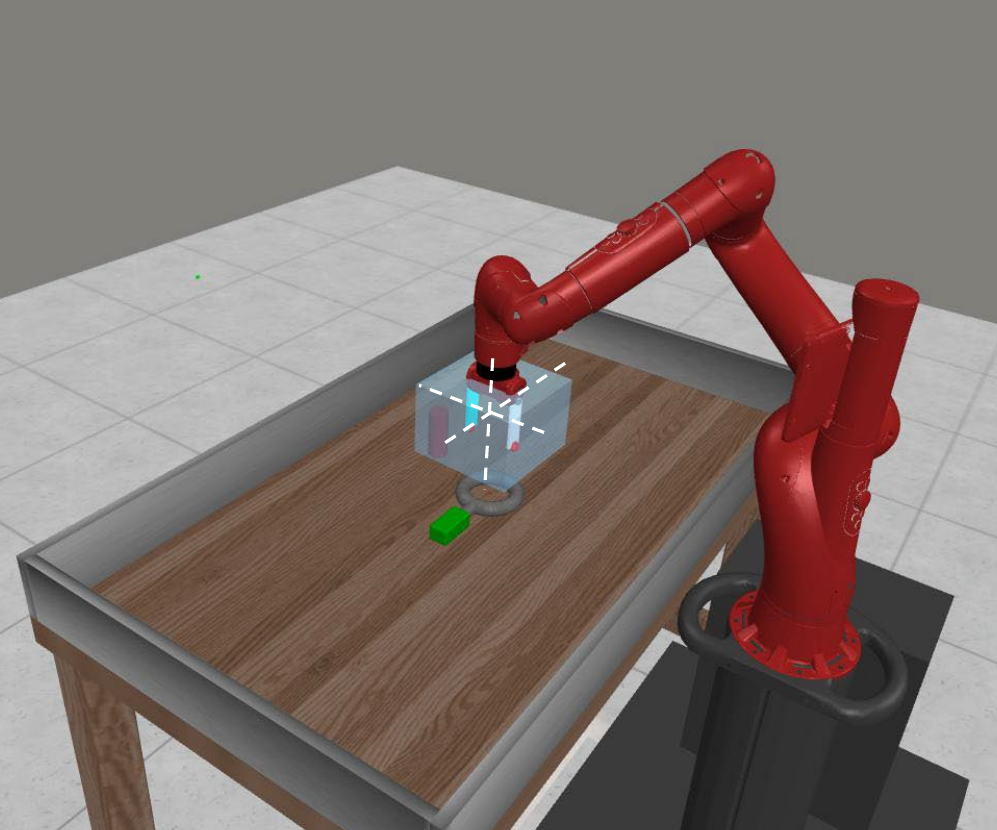}
        \caption{Assembly Env}
    \end{subfigure}
    \hfill
    \begin{subfigure}{0.32\textwidth}
        \includegraphics[width=\linewidth]{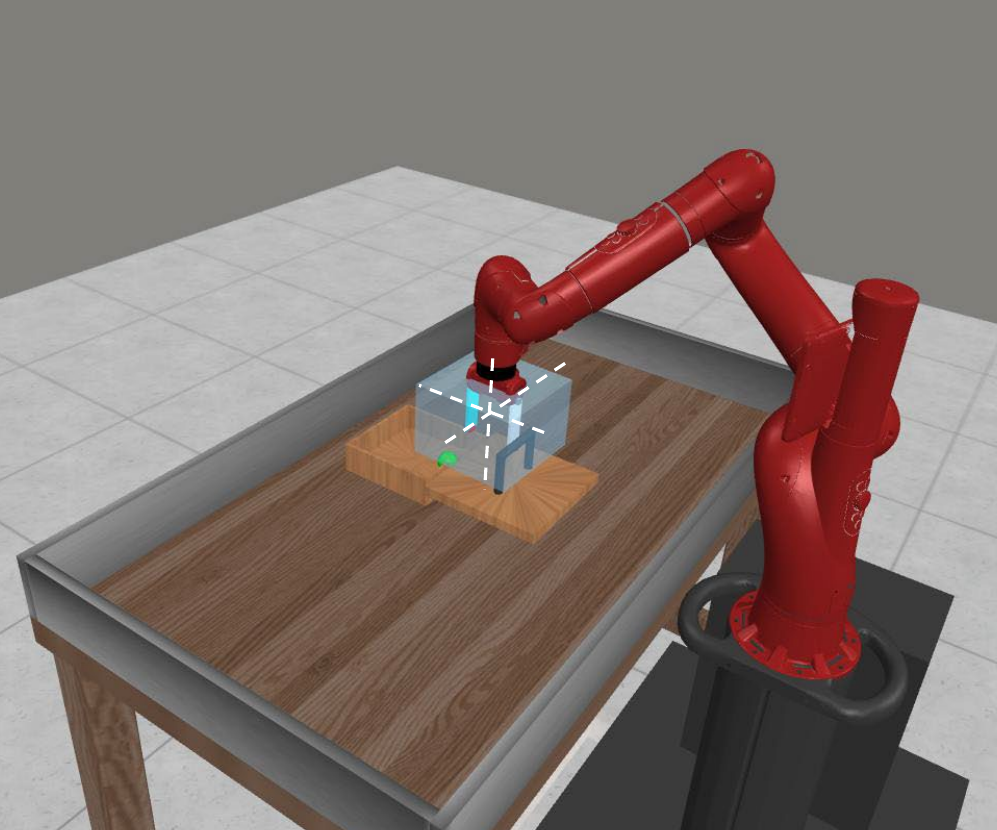}
        \caption{BoxClose Env}
    \end{subfigure}
    \hfill
    \begin{subfigure}{0.32\textwidth}
        \includegraphics[width=\linewidth]{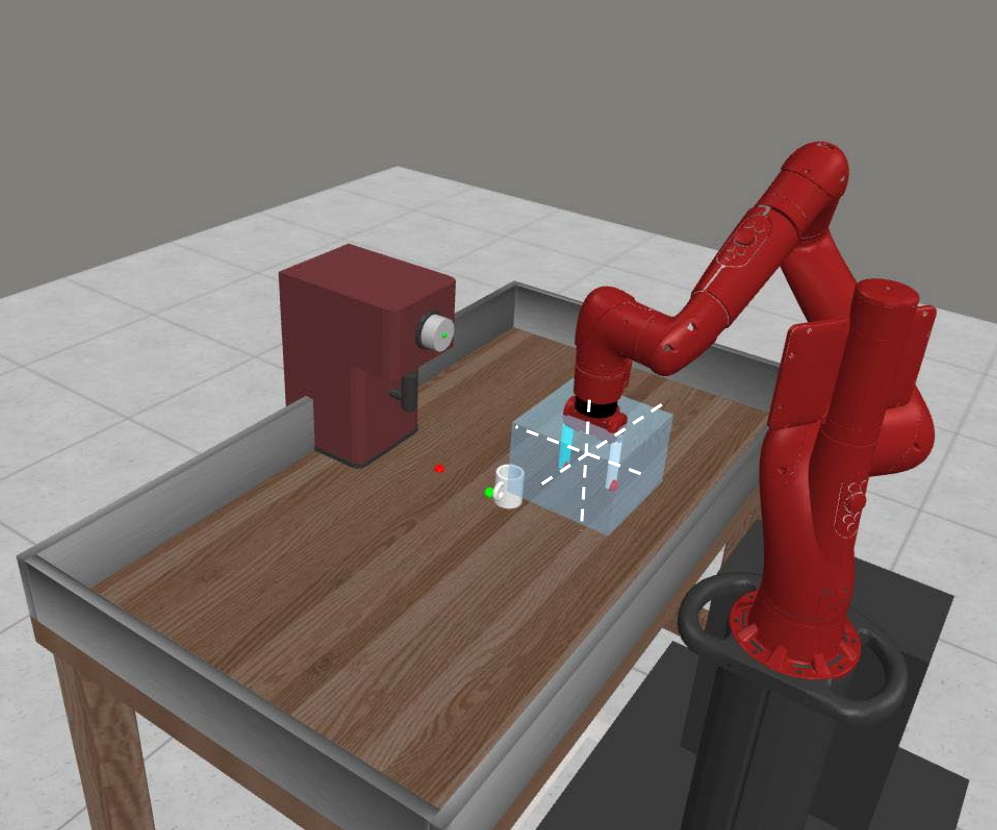}
        \caption{Coffeepush Env}
    \end{subfigure}

    \caption{This figure displays the evaluation setups for three MetaWorld tasks: Assembly, BoxClose, and CoffeePush, with randomized initial gripper positions. Each subfigure illustrates how the gripper's starting position varies in these environments, testing \NAVINACT{}'s ability to adapt to different initial conditions. 
    \label{gripper_rand_fig}
}
% \vspace{-10pt}
\end{figure}
\noindent \textbf{Can we predict the modes and waypoints accurately?}

\noindent\textbf{ModeNet} is trained on a dataset comprising 1,500 environmental camera images for simulation. The ground truth labels were determined by a distance threshold between the end-effector and the object, effectively deciding whether the robot should be in a motion-planning mode (moving towards an object) or an interaction mode (engaging directly with an object).  Data augmentation techniques like rotation and contrast adjustment were applied during training. The network utilizes cross-entropy loss and is optimized with Adam. The network's performance is evaluated using metrics such as accuracy, precision, recall, and F1 score. ModeNet achieved an accuracy score of 0.89, an F1 score of 0.85, a recall score of 0.9, and a precision score of 0.8, demonstrating its effectiveness in accurately classifying operational modes for robotic tasks.
\begin{figure}[h]
    \centering
    \includegraphics[width=\linewidth]{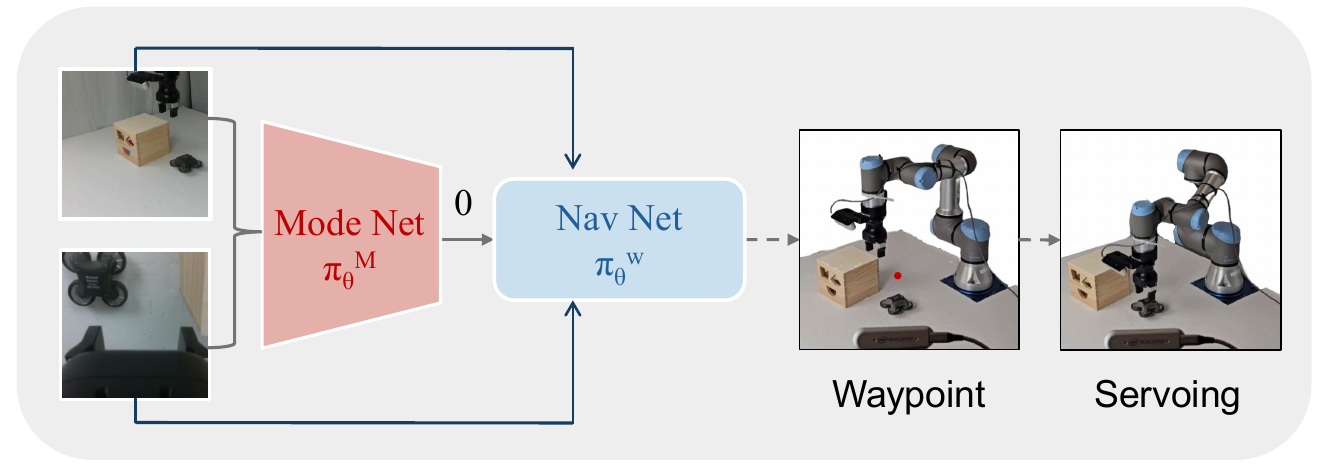}
    \caption{Sequence showing ModeNet predicting mode 0, followed by NavNet guiding the robotic arm to the designated waypoint.}
    % \vspace{-30pt}
    \label{navnet_demo}
\end{figure} 
\subsubsection{NavNet}
is trained on the same dataset of 1,500 environmental camera images as ModeNet. The ground truth for NavNet involves waypoints defined as positions 2 cm above the objects, guiding the robot on where to move next. Like ModeNet, data augmentation techniques such as rotation and contrast adjustments were used to enhance training robustness. NavNet employs Mean Squared Error (MSE) loss and is optimized with the Adam optimizer. It achieved an impressive accuracy score of 0.93, indicating high effectiveness in predicting accurate motion-planning waypoints based on visual inputs. Fig. \ref{navnet_demo} shows basic demo of \NAVINACT{} in motion-planning mode.
\subsubsection{ModeNet and NavNet for real-world experiments}

For real-world experiments, ModeNet and NavNet were trained using a more extensive dataset of 6200 images, incorporating both wrist and environment camera images.  The dataset captures a wider range of object perspectives, lighting conditions, and spatial contexts. This expansion improves the robustness and accuracy in real-world applications.

\subsection{Simulation Experiments}

Our evaluation suite includes three tasks from MetaWorld \cite{metaworld}, each using sparse 0/1 task completion rewards at the end of each episode. These tasks cover medium and hard difficulty levels as categorized in \cite{task_level}. 

The specific environments used are assembly-v2, box-close-v2, and coffee-push-v2. Given the absence of human demonstrations in MetaWorld, we utilized the demonstration datasets generated by IBRL, which were created using scripted expert policies from \cite{metaworld}. Although these tasks are relatively simple, they are sufficient to differentiate between stronger methods, especially in terms of sample efficiency and generalizability.

\noindent\textbf{Implementation of \NAVINACT{} and Baselines:}

\noindent \NAVINACT{} employs Behavior Cloning (BC) for imitation learning (IL) and Twin Delayed Deep Deterministic Policy Gradient (TD3) for reinforcement learning (RL). A ResNet-18 vision encoder is used for BC, and random-shift image augmentation is applied to enhance RL performance. \NAVINACT{} is benchmarked against three baselines: IBRL \cite{ibrl}, RL-MN (RL with ModeNet and NavNet), and standard RL.

\textbf{IBRL} uses a pre-trained IL policy, \(\mu_\psi\), to aid exploration during RL and to assist in target value estimation during Temporal Difference (TD) learning. Likewise, \NAVINACT{} also uses IL policy within the InteractNet, that speeds up RL during object interaction. RL with ModeNet and NavNet, which we refer as \textbf{RL-MN}, incorporates ModeNet and NavNet to accelerate exploration by predicting waypoints, reducing the number of exploration steps needed in RL. Standard \textbf{RL} utilizes TD3 but does not use any IL or hierarchical mode prediction components.
\subsection{Results} 
\begin{figure}[H]
    \centering
    \begin{subfigure}{0.32\textwidth}
        % \hspace{0.08\textwidth}
        % \raisebox{0.15\height}{\includegraphics[width=0.95\textwidth, height=0.15\textheight]
        % {isrr2024_format/templates/figures/assembly_env_render.pdf}}
        \includegraphics[width=\linewidth]{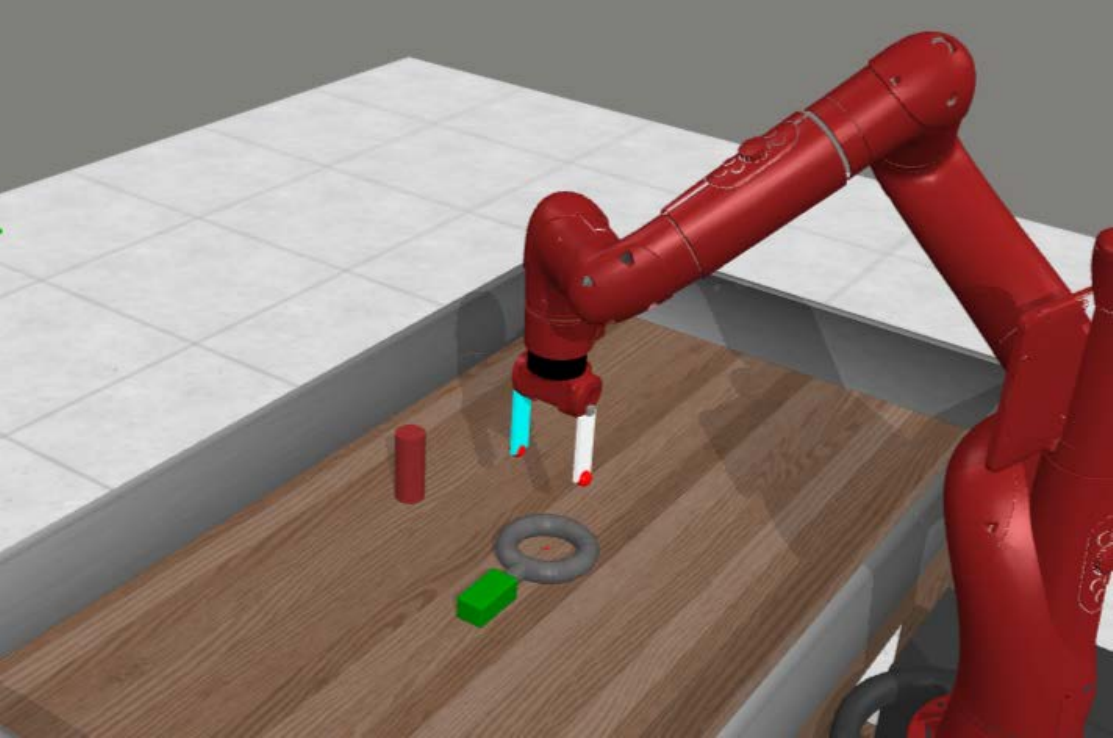}
        \caption{Assembly Env}
        \label{assembly_env}
    \end{subfigure}
    \hfill
    \begin{subfigure}{0.32\textwidth}
        % \hspace{0.08\textwidth}
        % \raisebox{0.15\height}{\includegraphics[width=0.95\textwidth, height=0.15\textheight]
        % {isrr2024_format/templates/figures/assembly_env_render.pdf}}
        \includegraphics[width=\linewidth]{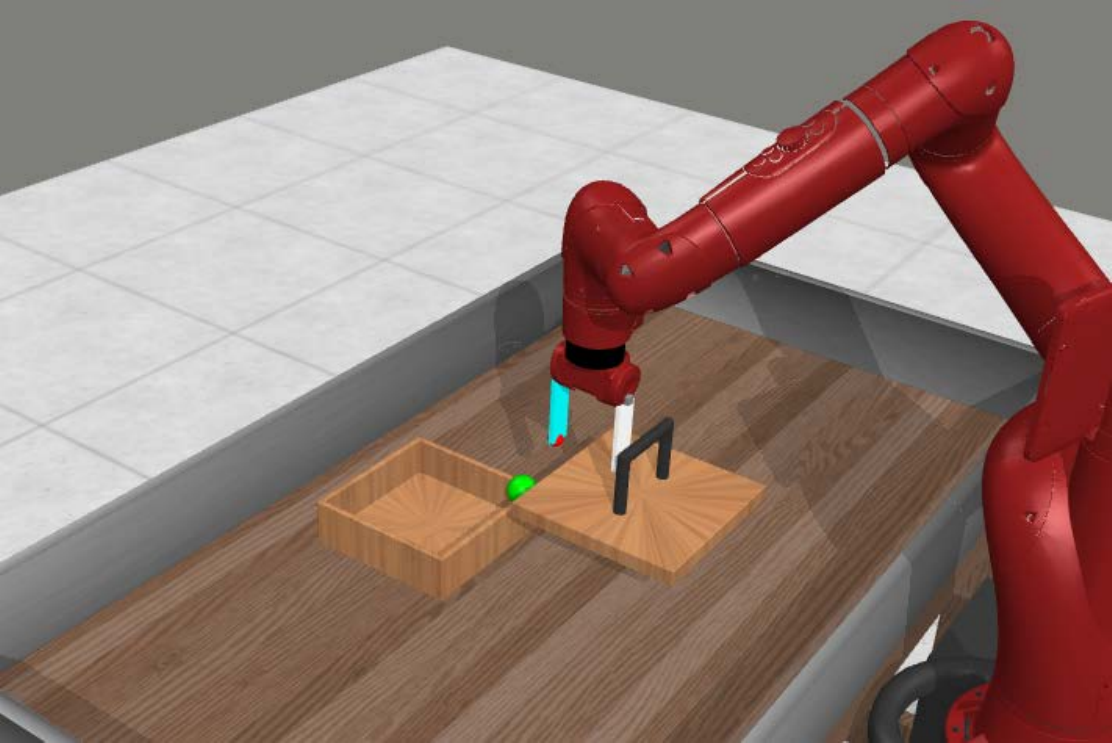}
        \caption{Boxclose Env}
        \label{boxclose_env}
    \end{subfigure}
    \hfill
    \begin{subfigure}{0.32\textwidth}
        % \hspace{0.08\textwidth}
        % \raisebox{0.15\height}{\includegraphics[width=0.95\textwidth, height=0.15\textheight]
        % {isrr2024_format/templates/figures/assembly_env_render.pdf}}
        \includegraphics[width=\linewidth]{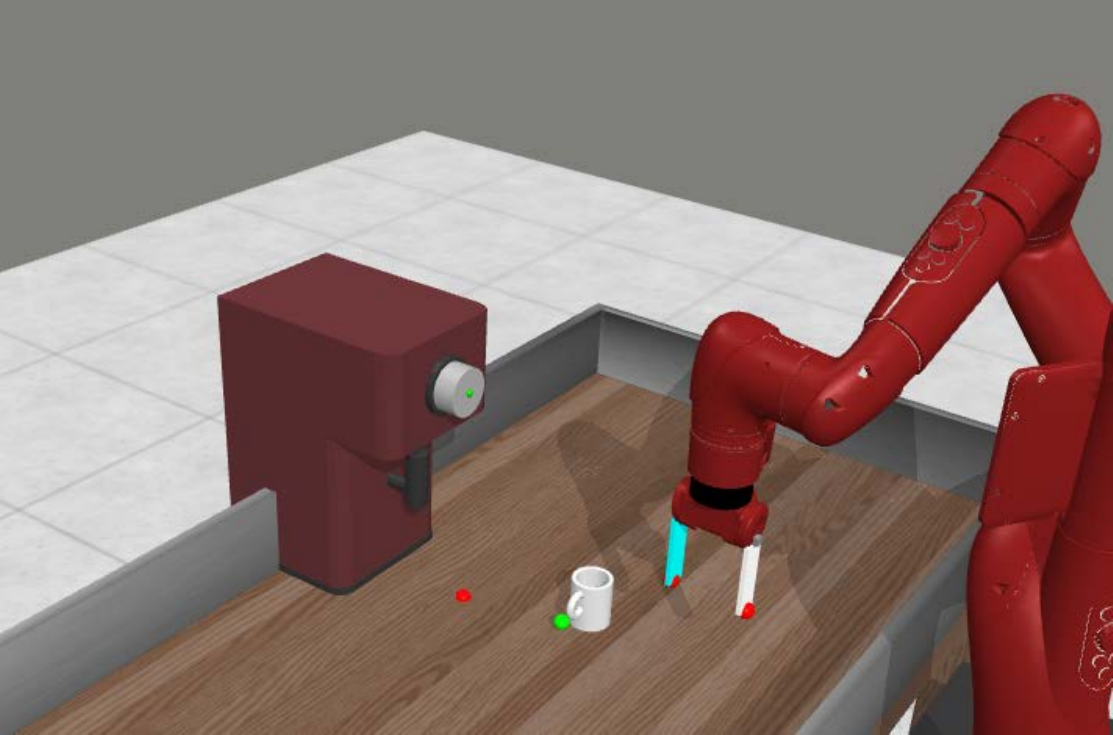}
        \caption{CoffeePush Env}
        \label{coffeepush_env}
    \end{subfigure}

    \begin{subfigure}{0.32\textwidth}
        \includegraphics[width=\linewidth]{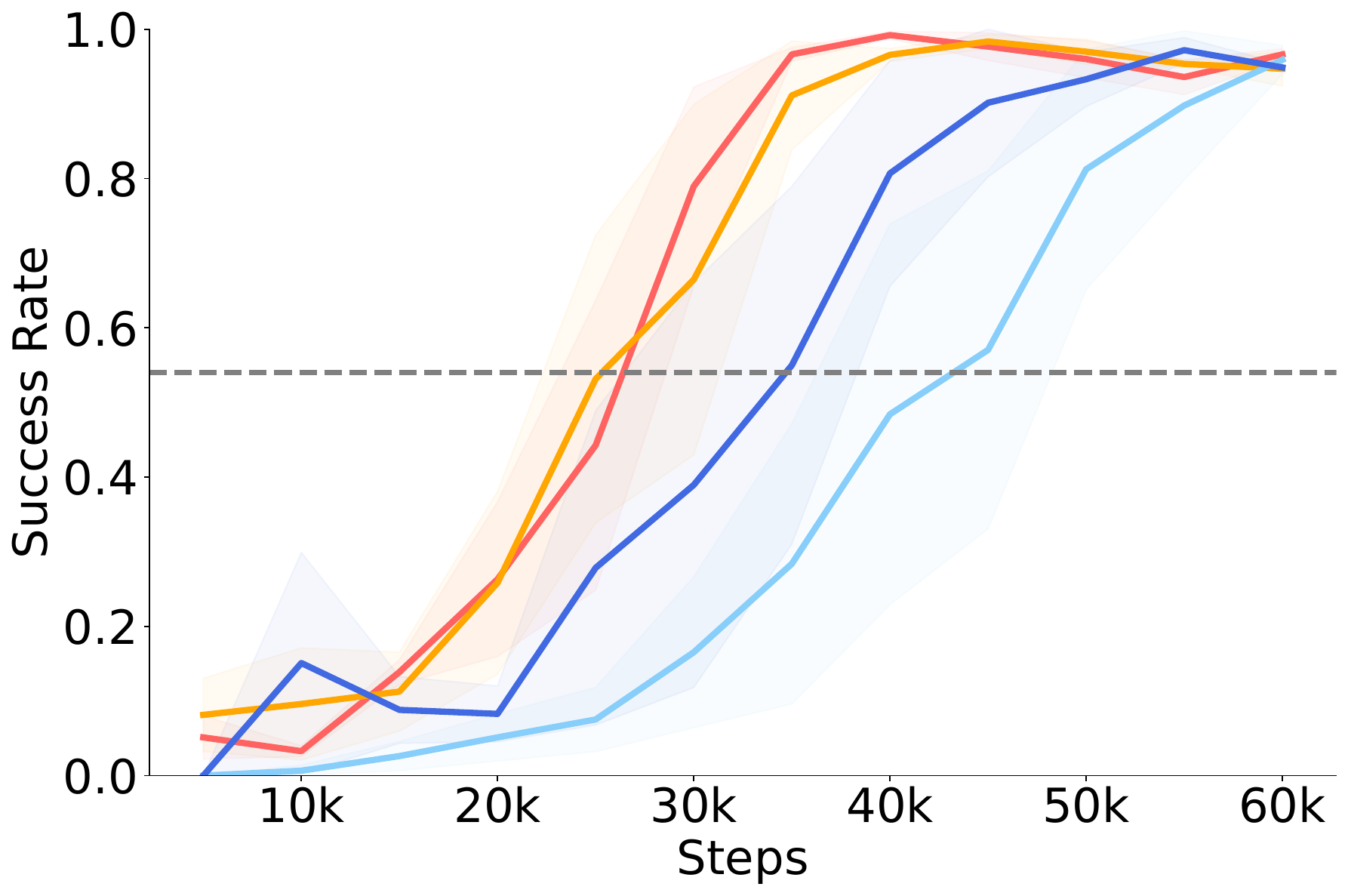}
        \caption{Training with random initial Objects position.}
        \label{assembly_train}
    \end{subfigure}
    \begin{subfigure}{0.32\textwidth}
        \includegraphics[width=\linewidth]{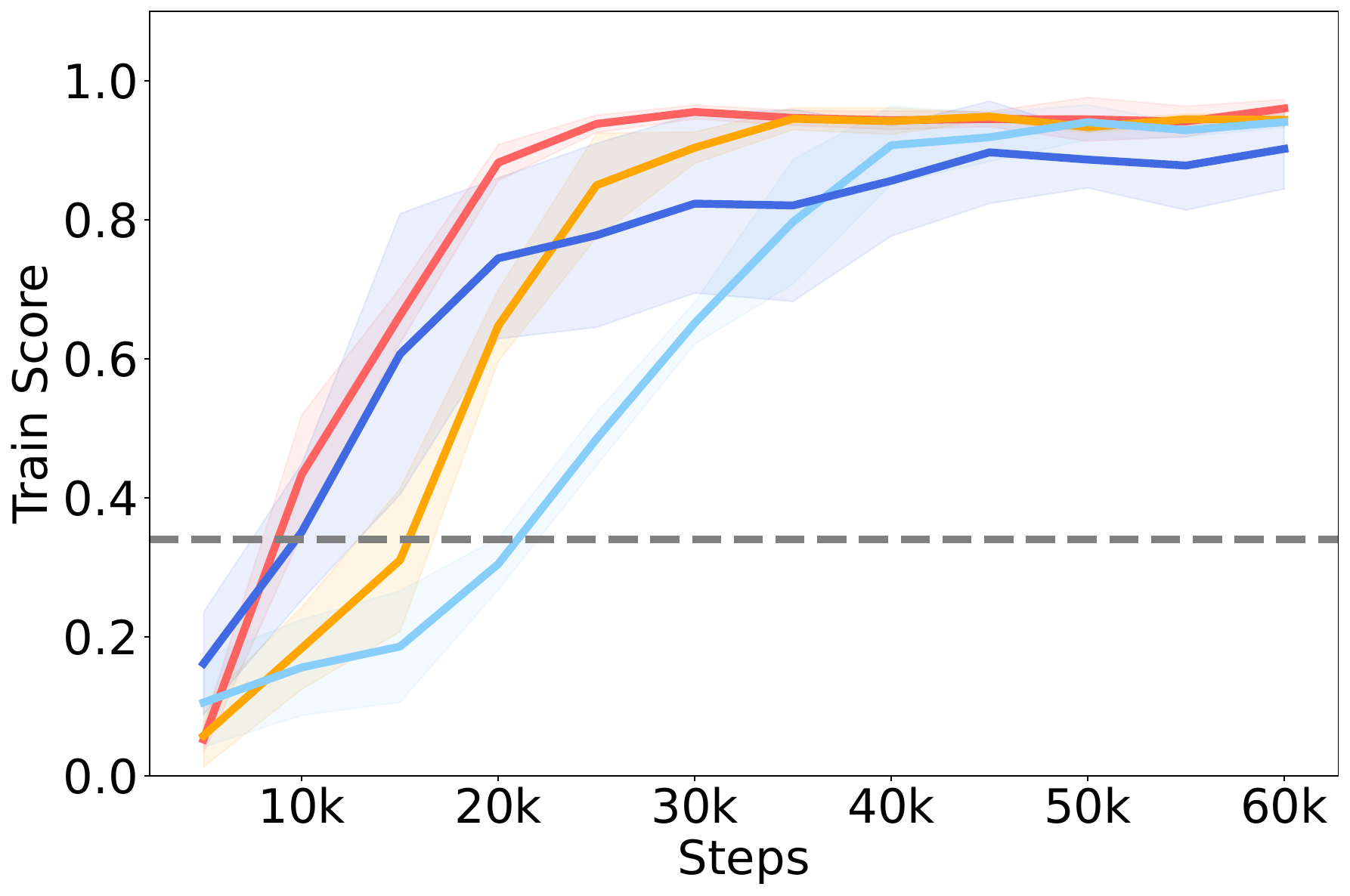}
        \caption{Training with random initial Objects position.}
        \label{boxclose_train}
    \end{subfigure}
    \begin{subfigure}{0.32\textwidth}
        \includegraphics[width=\linewidth]{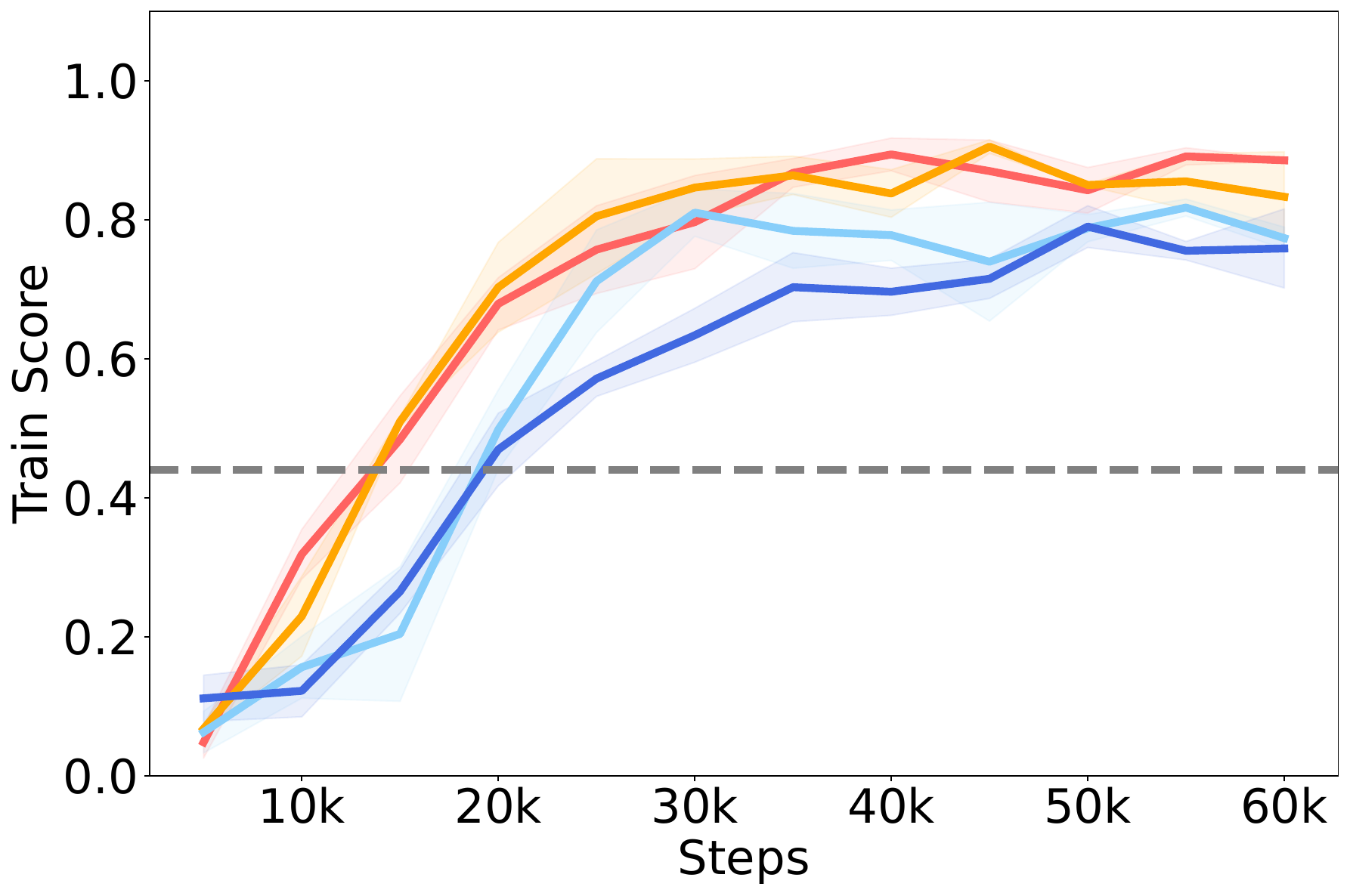}
        \caption{Training with random initial Objects position.}
        \label{coffeepush_train}
    \end{subfigure}

    \begin{subfigure}{0.32\textwidth}
        \includegraphics[width=\linewidth]{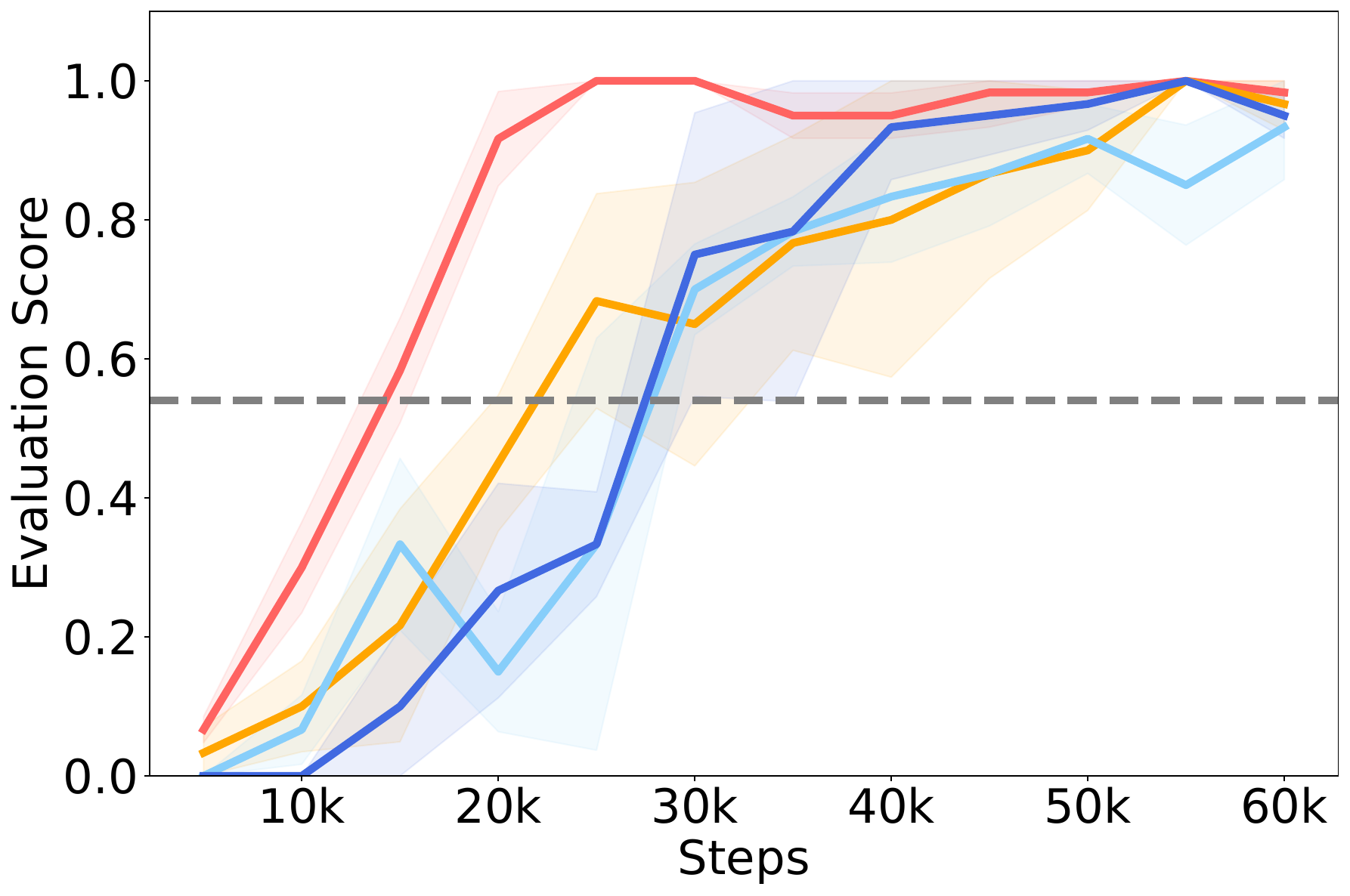}
        \caption{Evaluation with random initial Objects position.}
        \label{assembly_eval}
    \end{subfigure}
    \begin{subfigure}{0.32\textwidth}
        \includegraphics[width=\linewidth]{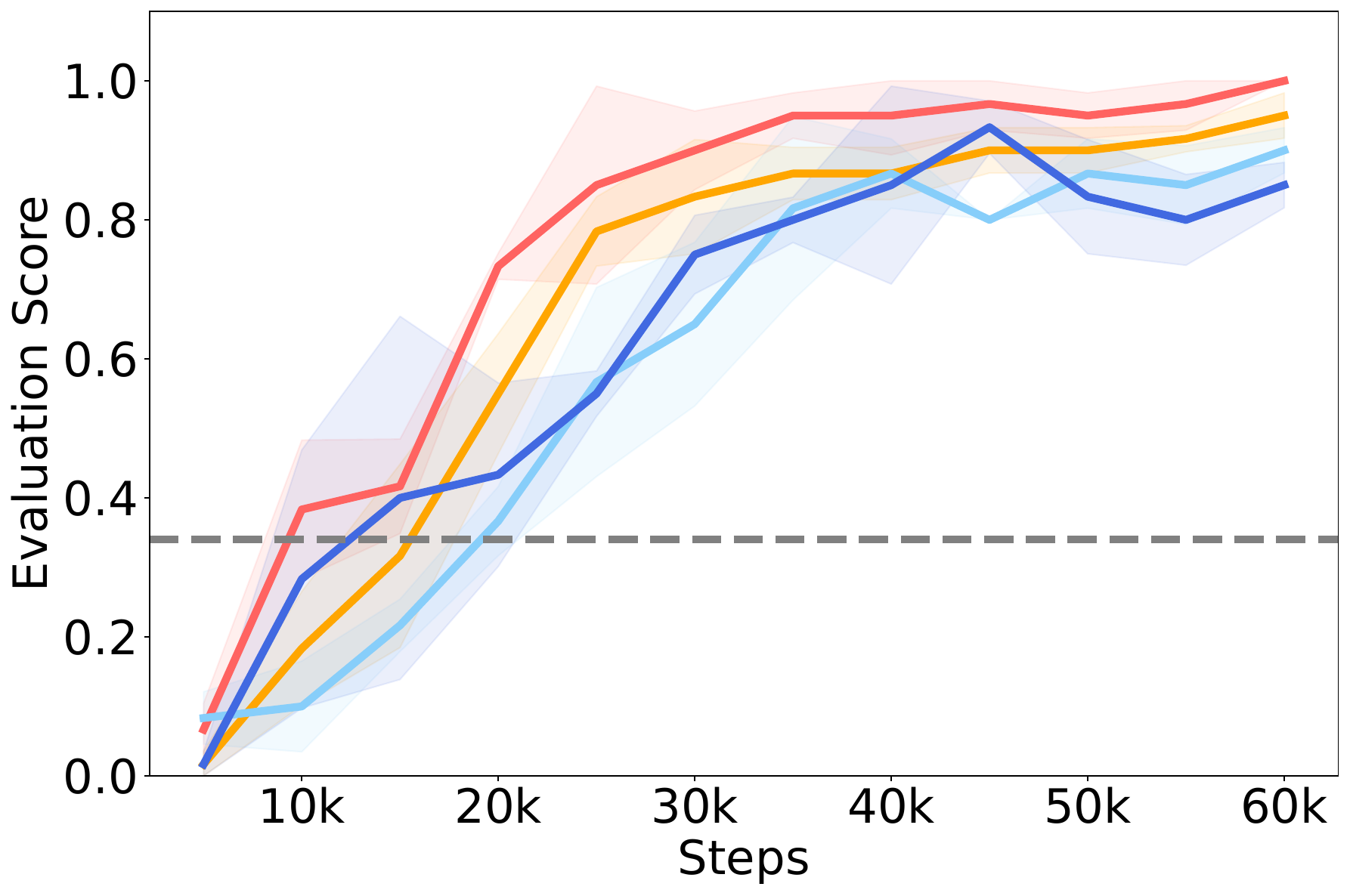}
        \caption{Evaluation with random initial Objects position.}
        \label{boxclose_eval}
    \end{subfigure}
    \begin{subfigure}{0.32\textwidth}
        \includegraphics[width=\linewidth]{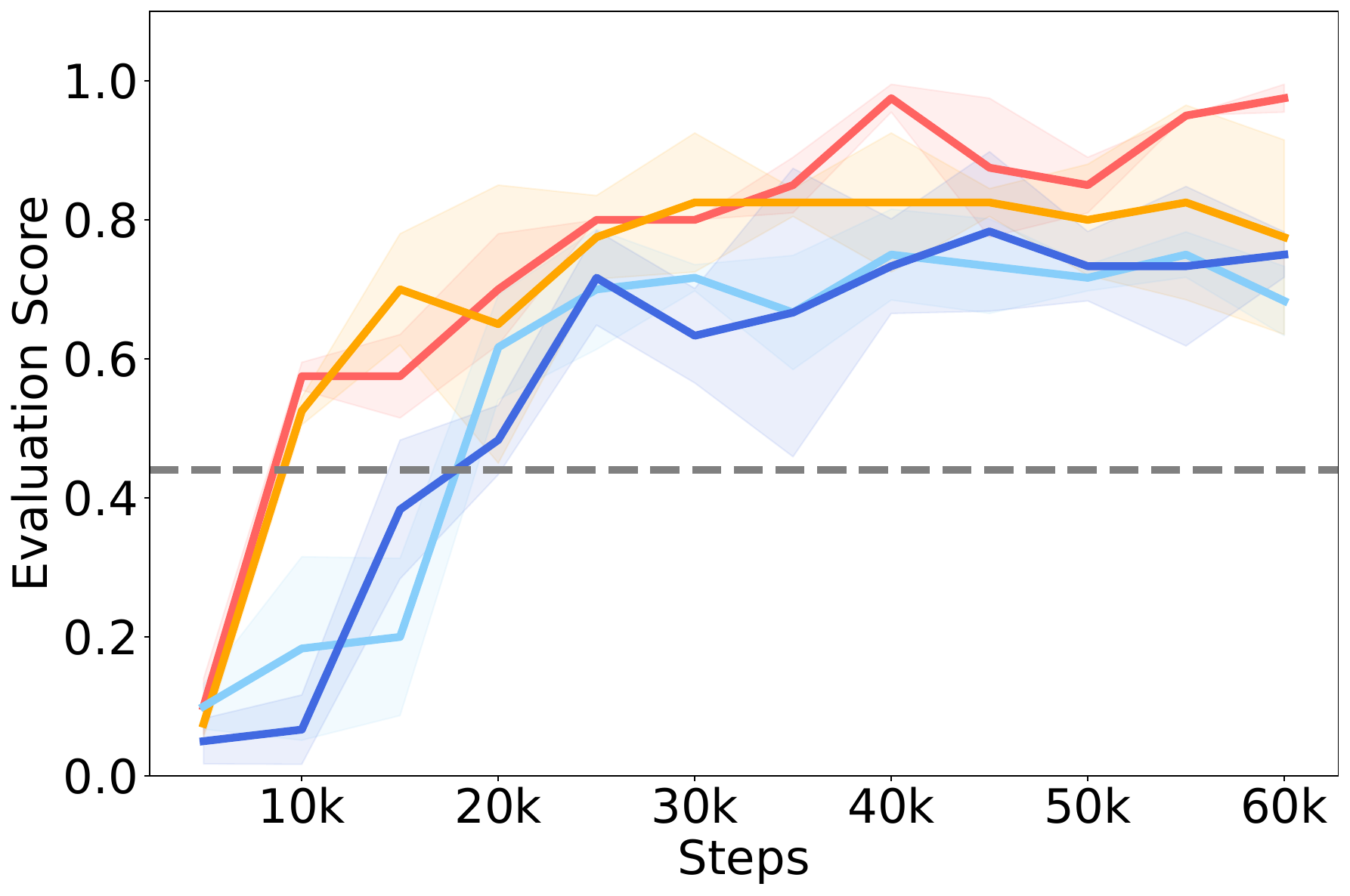}
        \caption{Evaluation with random initial Objects position.}
        \label{coffeepush_eval}
    \end{subfigure}

    \begin{subfigure}{0.32\textwidth}
        \includegraphics[width=\linewidth]{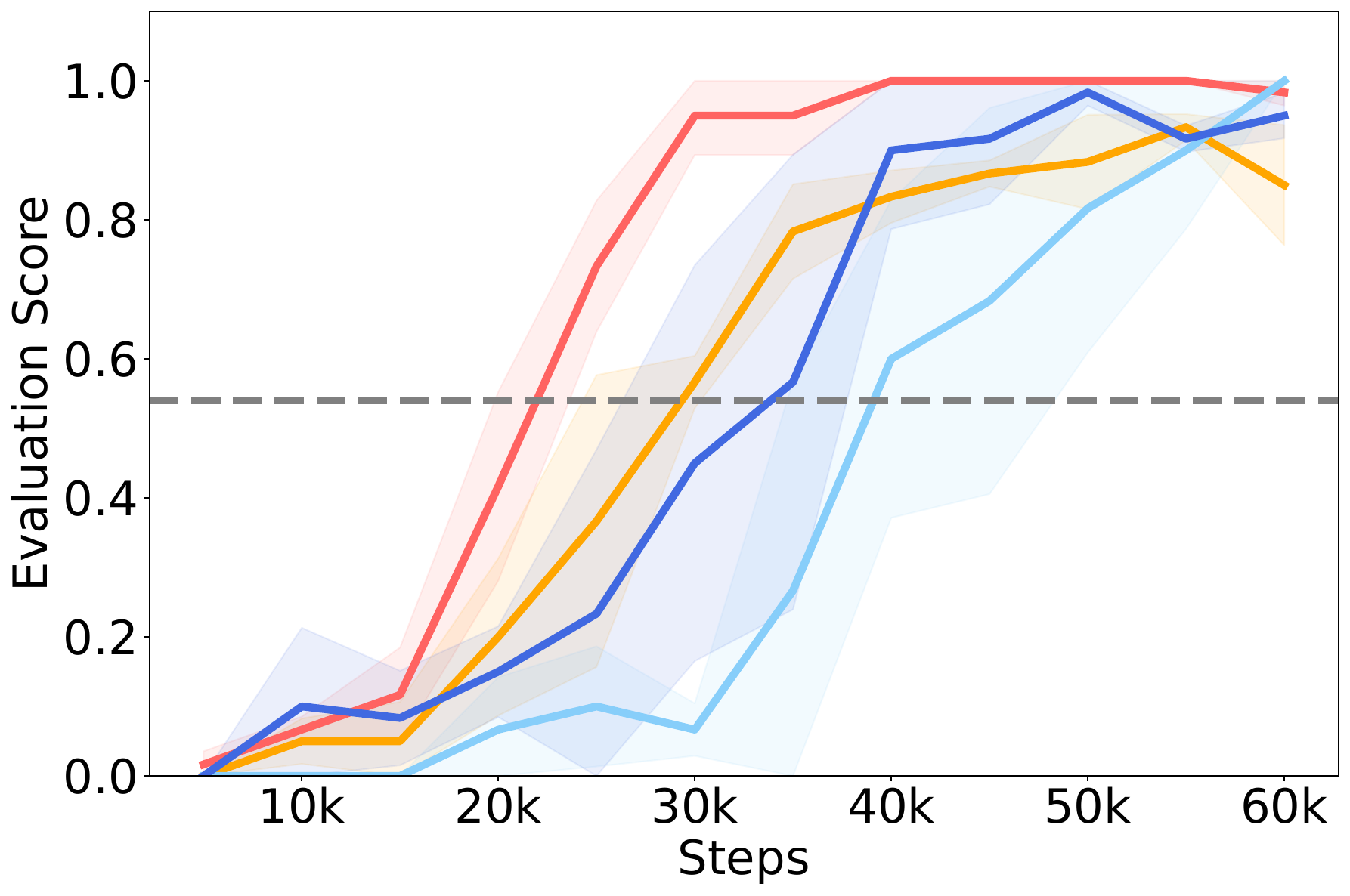}
        \caption{Evaluation with random initial Objects and gripper position.}
        \label{assembly_eval_rand}
    \end{subfigure}
    \hfill
    \begin{subfigure}{0.32\textwidth}
        \includegraphics[width=\linewidth]{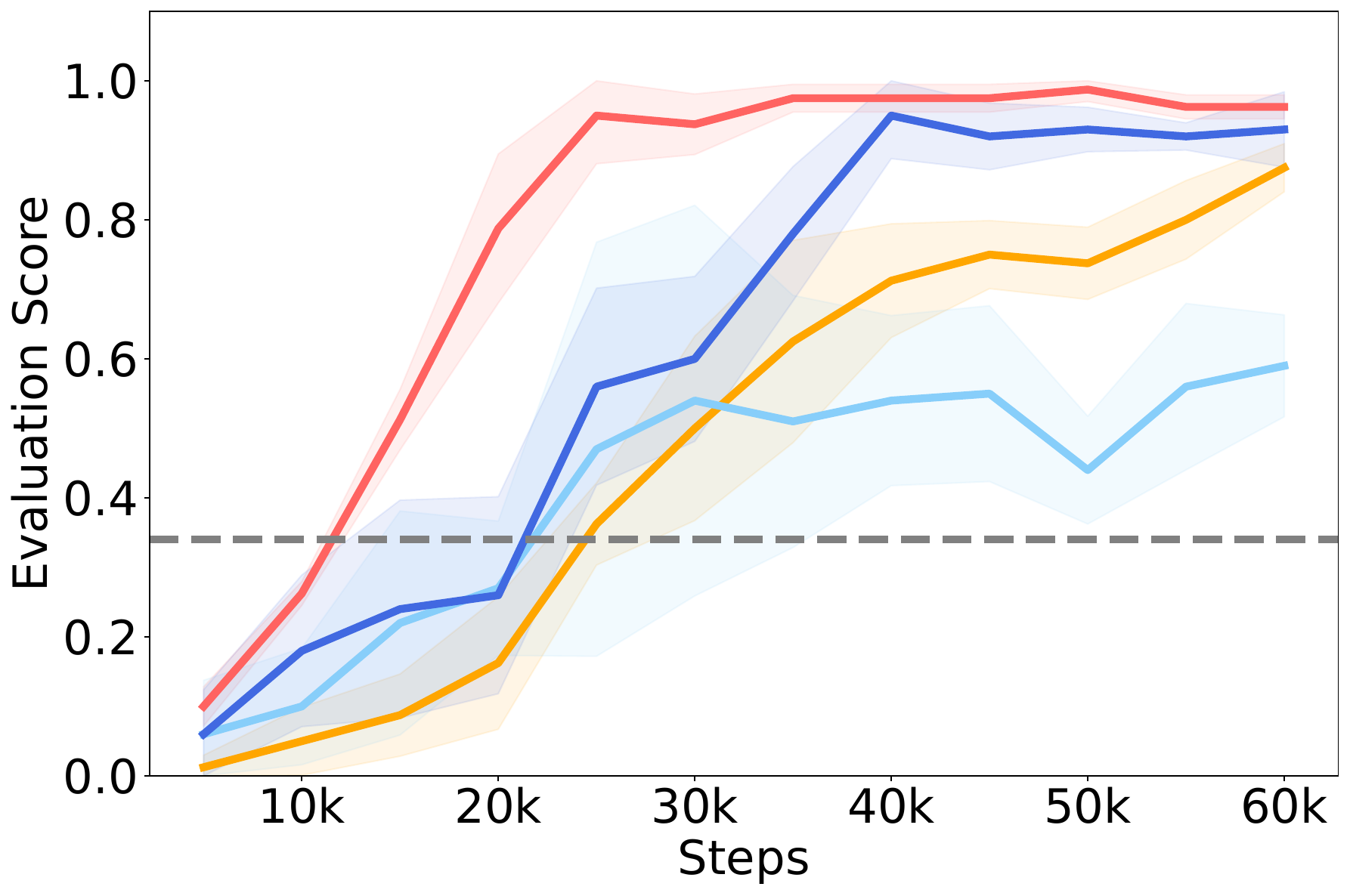}
        \caption{Evaluation with random initial Objects and gripper position.}
        \label{boxclose_eval_rand}
    \end{subfigure}
    \hfill
    \begin{subfigure}{0.32\textwidth}
        \includegraphics[width=\linewidth]{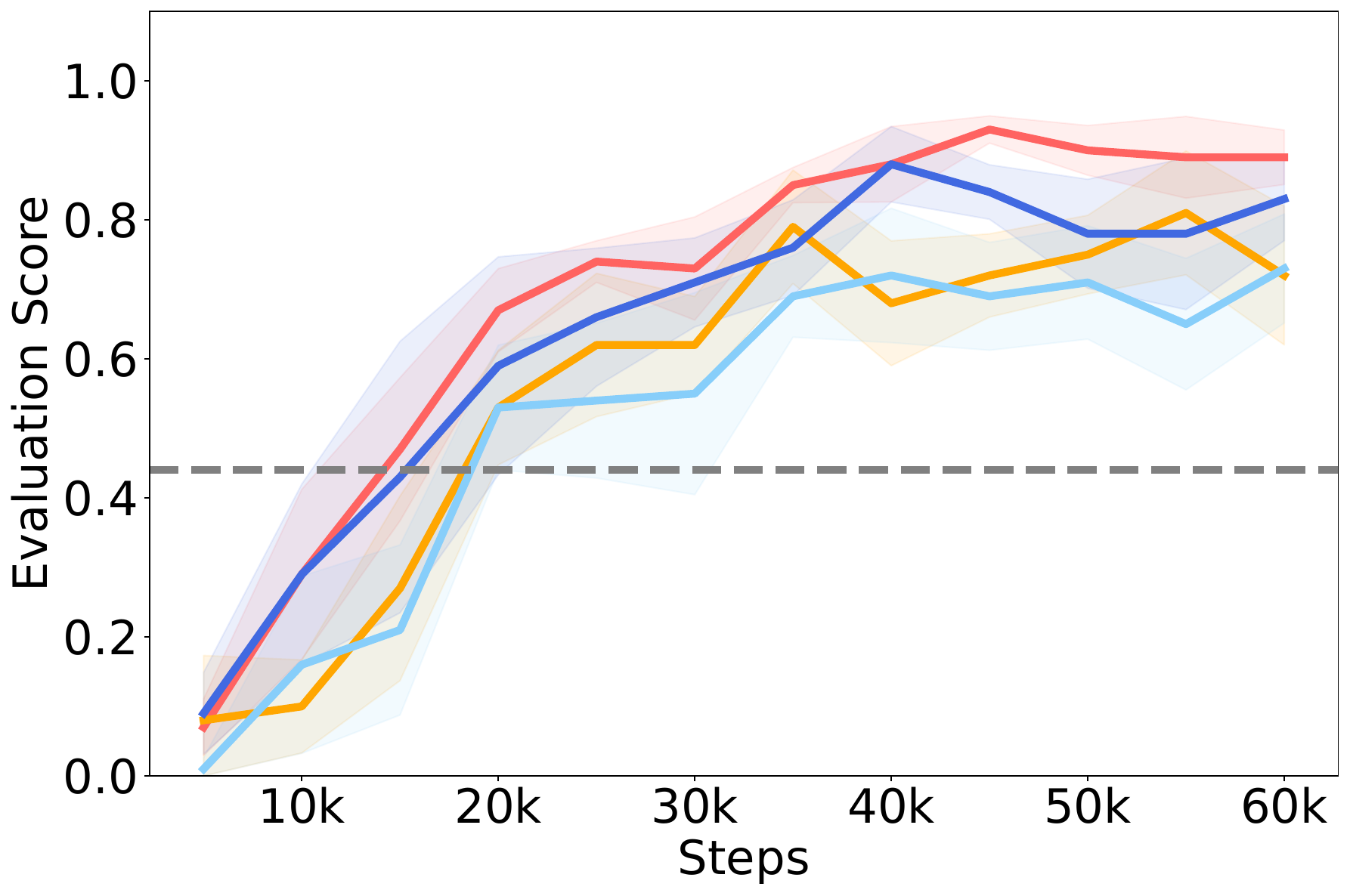}
        \caption{Evaluation with random initial Objects and gripper position.}
        \label{coffeepush_eval_rand}
    \end{subfigure}
    
    \begin{subfigure}{0.7\textwidth}
        \includegraphics[width=\linewidth]{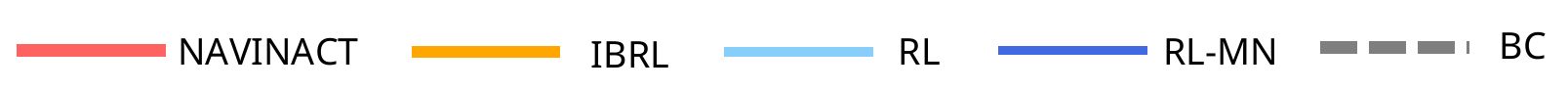}
    \end{subfigure} 
    \caption{Performance Evaluation of \NAVINACT{} Across MetaWorld Tasks. This figure presents the training and evaluation results for three challenging MetaWorld environments: (a) Assembly, (b) BoxClose, and (c) CoffeePush. The top row depicts these environments in their initial configurations. The middle row (d), (e) and (f) shows the training performance of \NAVINACT{} under conditions where object positions are randomized within environments (a), (b), and (c). The bottom two rows illustrate the evaluation performance under different randomized scenarios: varying initial object positions and combined randomization of both object and gripper positions for each environment. The results demonstrate \NAVINACT{}'s robustness and adaptability to varying task complexities, maintaining high performance across different training and evaluation conditions.}
    % \vspace{-7pt}
    \label{sim_results}
\end{figure}

% \vspace{-20pt}
% \textbf{NAVINACT matches or exceeds baselines in Meta-World.} 
Following the training of \NAVINACT{} and the baselines, we evaluate its performance on unseen configuration of object positions in the environments and compare it against baseline methods. Performance is assessed using training and evaluation scores, which indicate sample efficiency and generalizability.  Fig. \ref{sim_results} illustrates the comparative results for all the metaworld environments, including training and evaluation scores for various initial positions. 

% \FloatBarrier % Ensures that the figure is placed before this barrier
Both \NAVINACT{} and IBRL utilize a trained BC policy, with \NAVINACT{} achieving a higher evaluation score of 0.38 in the assembly environment, while RL-MN and RL do not incorporate this policy. Our evaluation is focused on 3 key metrics --- training score for the simulation tasks with randomization with respect to object position, evaluation score with randomization with respect to object position, and evaluation score with randomization with respect to random initial gripper (as shown in Fig. \ref{gripper_rand_fig}) as well as object positions.
The performance of \NAVINACT{}, IBRL, RL-MN, and standard RL is compared across the Meta-World environments over 60k timesteps and 7 seeds.

\NAVINACT{} outperforms baselines across all tasks in terms of both sample efficiency
and final performance, solving all tasks within 30K samples.

\subsubsection{Does \NAVINACT{} Lead to Better Sample Efficiency?} 
To assess sample efficiency, we compare the training curves of \NAVINACT{} with those of IBRL, RL-MN, and RL across various MetaWorld tasks. These tasks represent different levels of complexity, categorized into medium, hard, and very hard tiers. \NAVINACT{} demonstrates better sample efficiency, as evidenced by faster convergence and higher success rates with fewer interaction steps as shown in Fig. \ref{assembly_train},\ref{boxclose_train} and \ref{coffeepush_train}. This improvement is attributed to \NAVINACT{}'s mode prediction, which effectively reduces task complexity by narrowing down the action space at each step. \NAVINACT{} consistently achieves an impressive success rate of 85±5\% during training and learns quickly within 30k sample steps.

\begin{wrapfigure}{r}{0.5\textwidth}
    \centering
    \includegraphics[width=\linewidth]{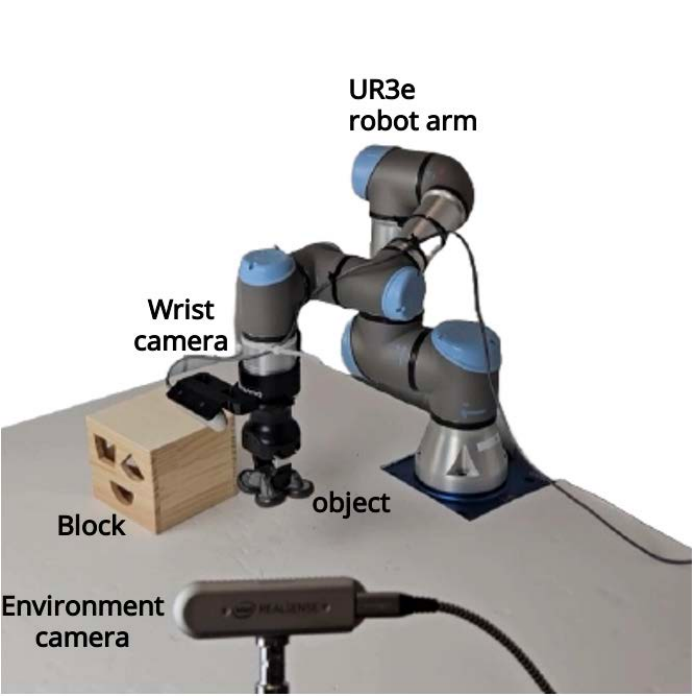}
    \caption{Experimental Setup}
     % \vspace{-10pt}
    \label{real_world_setup}
\end{wrapfigure}
\subsubsection{Does \NAVINACT{} Generalize to Unseen Conditions?}
To evaluate generalization, we test \NAVINACT{} in conditions not included in the training data, such as variations in object positions, and initial gripper position. \NAVINACT{} shows strong generalization capabilities, significantly outperforming IBRL and pure RL methods in unseen scenarios as indicated in Fig. \ref{assembly_eval},\ref{boxclose_eval},\ref{coffeepush_eval},\ref{assembly_eval_rand},\ref{boxclose_eval_rand} and \ref{coffeepush_eval_rand}. The mode prediction network plays a crucial role in adapting to new conditions by dynamically adjusting the operational mode, which is not achievable by methods solely relying on IL. Our algorithm maintains a nearly equivalent rate of 80±5\% in evaluation phases across unseen configurations of object position in the environments. 
We see that, \NAVINACT{} matches or surpasses other baselines in Meta-World. 
Notably, as randomization increases, the performance of IBRL, RL-MN, and standard RL decreases, while \NAVINACT{} maintains consistent performance and demonstrates robustness to variability and changes in the environment.

\subsection{Real-world Experimental Setup}

\begin{figure}[h]
    \centering
    \begin{subfigure}{0.33\textwidth}
        % \hspace{0.08\textwidth}
        % \raisebox{0.15\height}{\includegraphics[width=0.95\textwidth, height=0.15\textheight]
        \includegraphics[width=\linewidth]{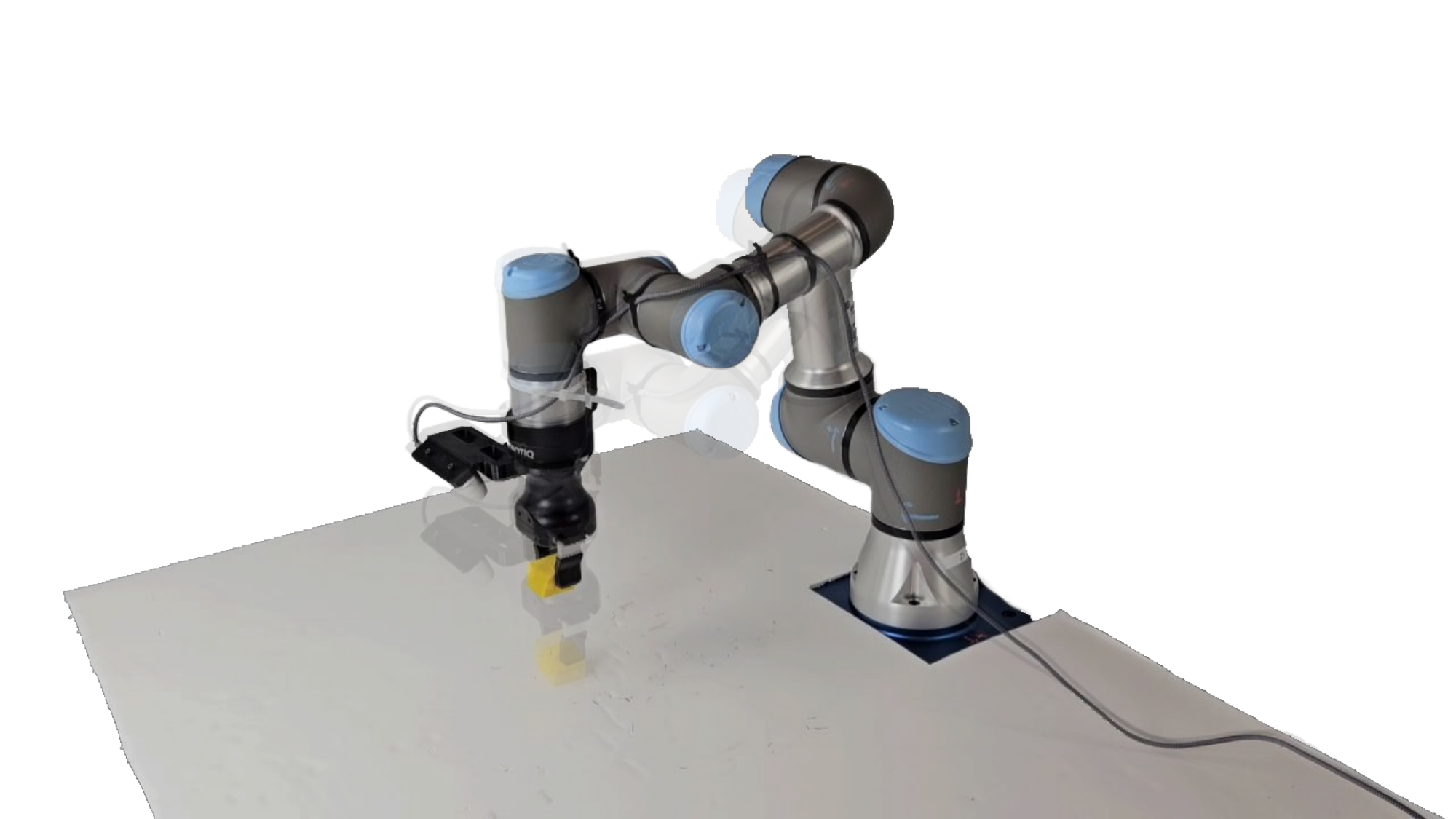}
        \caption{Lift}
    \label{lift}
    \end{subfigure}
    \hfill
    \begin{subfigure}{0.32\textwidth}
        \includegraphics[width=\linewidth]{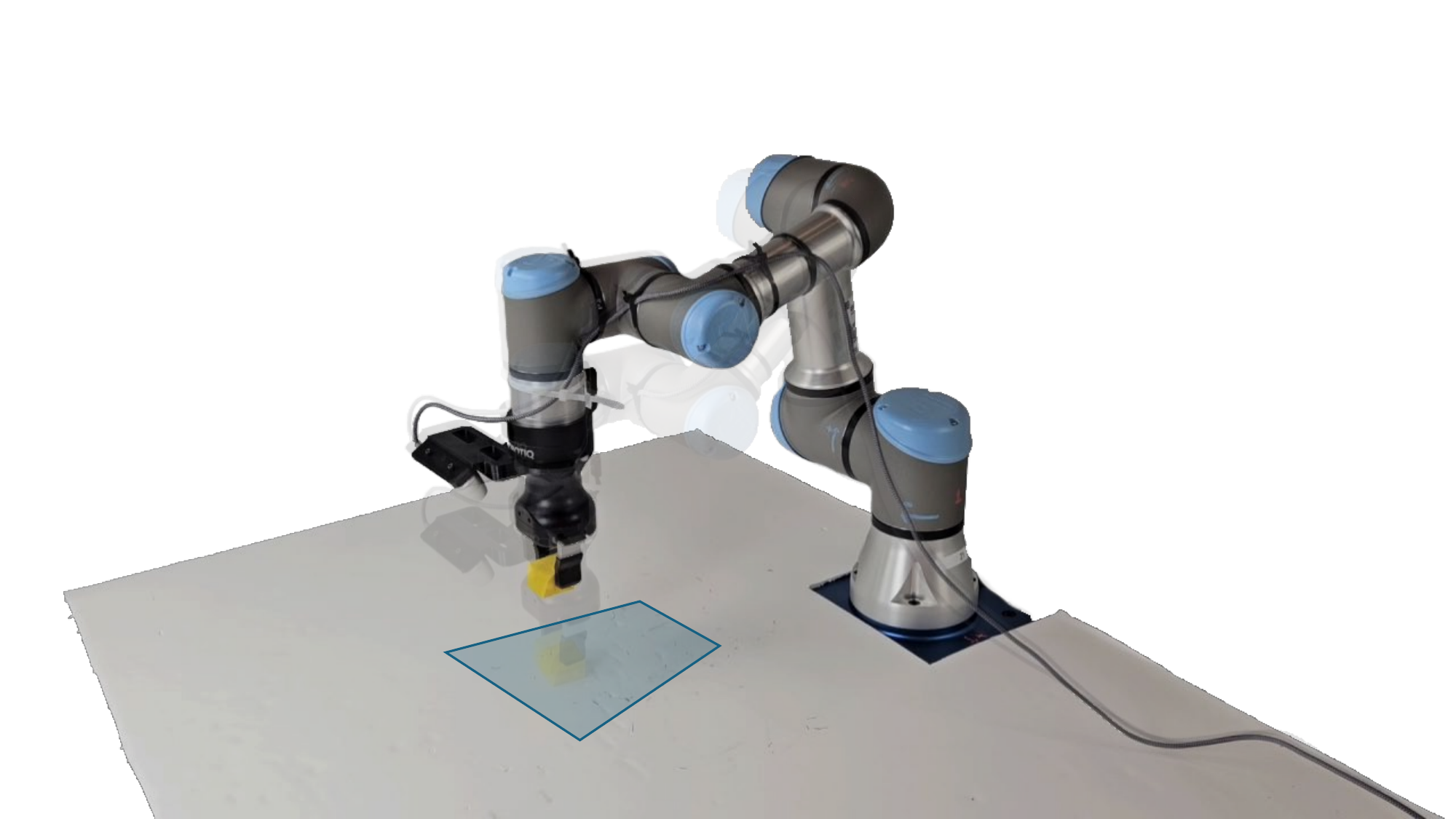}
        \caption{Lift with randomization}
    \label{lift_rand}
    \end{subfigure}
    \hfill
    \begin{subfigure}{0.33\textwidth}
        \includegraphics[width=\linewidth]{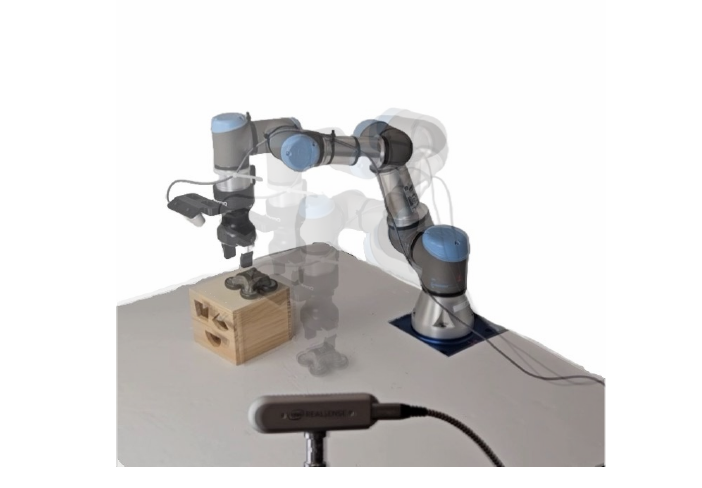}
        \caption{Pick and Place}
    \label{2s_env}
    \end{subfigure}

    \begin{subfigure}{0.33\textwidth}
        \includegraphics[width=\linewidth]{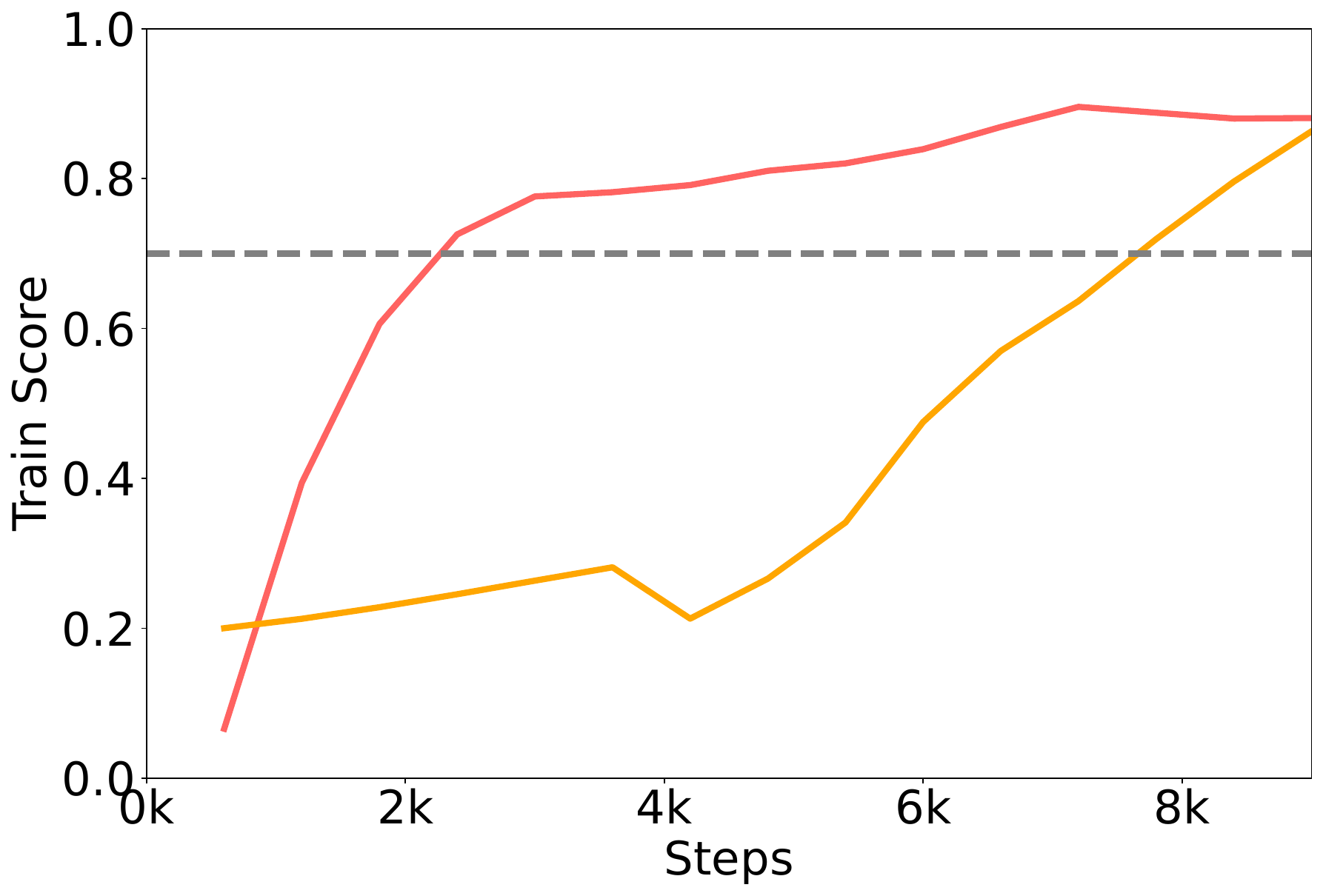}
        \caption{}

    \label{lift_result}
    \end{subfigure}
    \hfill
    \begin{subfigure}{0.32\textwidth}
        \includegraphics[width=\linewidth]{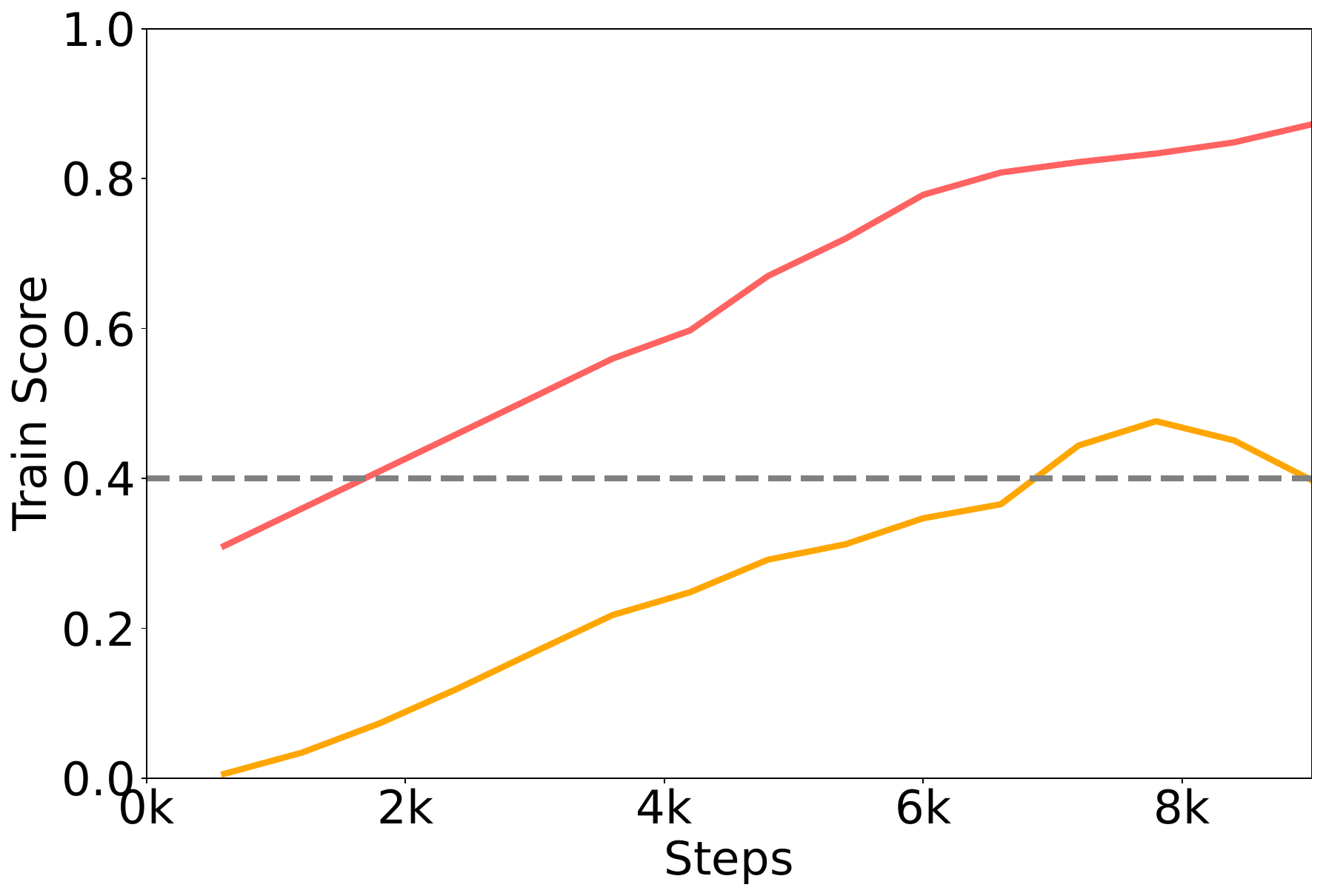}
        \caption{}

    \label {lift_rand_result}    
    \end{subfigure}
    \hfill
    \begin{subfigure}{0.33\textwidth}
        \includegraphics[width=\linewidth]{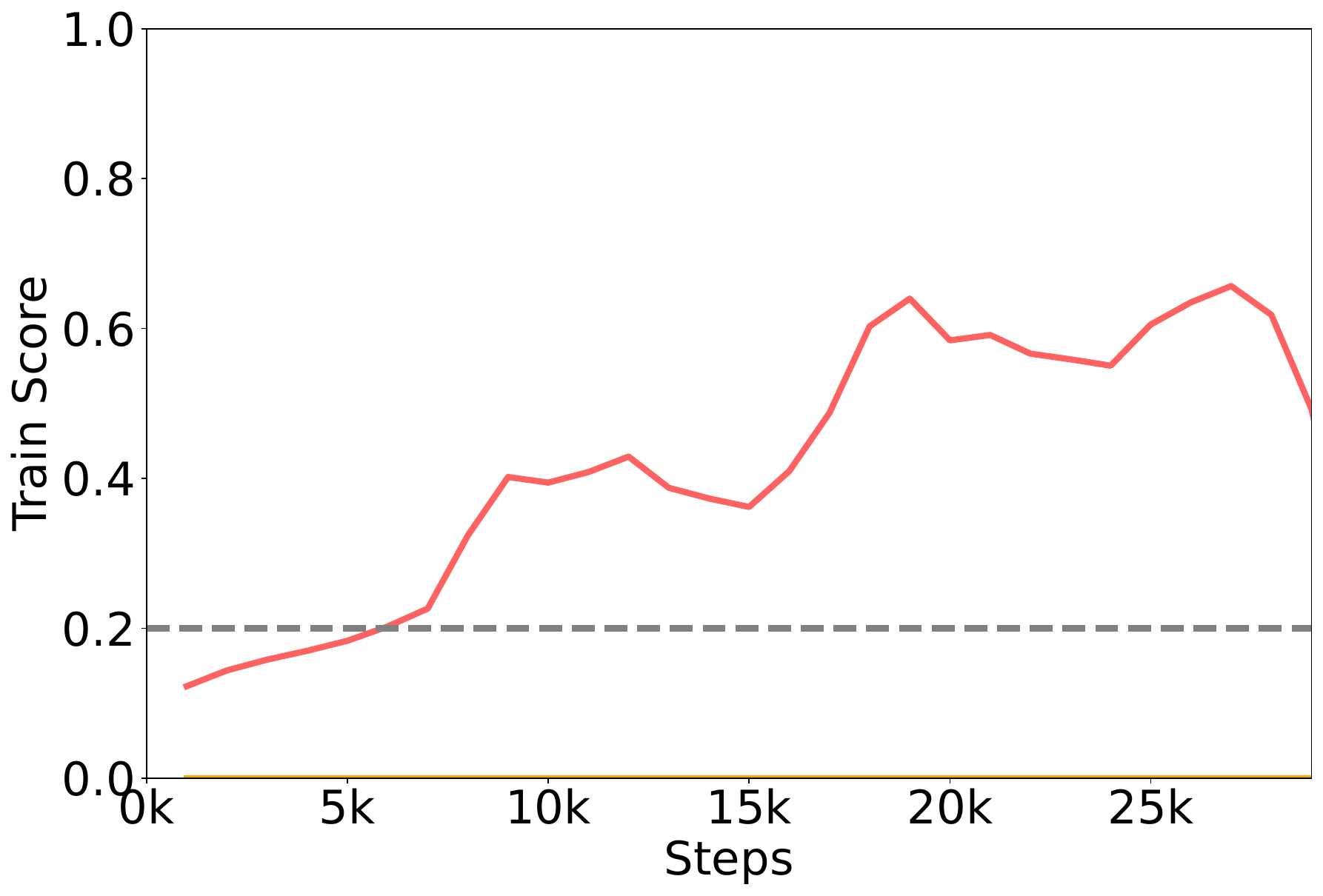}
        \caption{}

    \label{2s_result}    
    \end{subfigure}

    \begin{subfigure}{0.35\textwidth}
        \includegraphics[width=\linewidth]{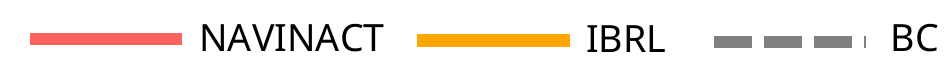}
    \end{subfigure} 

    \caption{\textbf{Top:} Illustrations of lift task and the variation in the
initialization of the task, left (fixed) right (randomized block position). Bottom: Performance evaluation on the lift task under both fixed and randomized conditions. The training curves depict the success rate over time, demonstrating \NAVINACT{}'s effective learning and adaptability in varying scenarios.}
% \vspace{-20pt}
\label{real_world_result}
\end{figure}

We design two tasks named Lift and pick-and-place as shown in Fig.\ref{lift}, \ref{lift_rand}, \ref{2s_env}. The
tasks use a UR3e robot. The robot is equipped
with a Robotiq hand-e gripper and a realsense camera mounted on the wrist. We also use another realsense camera as environment camera for the experiment as shown in Fig. \ref{real_world_setup}.
% \vspace{-40pt}

Actions are 4-dimensional
consisting of 3 dimensions for difference of end-effector position (deltas), under a Cartesian Delta controller and
1 dimension for absolute position of the gripper. Policies
run at 6 Hz. For each task, we collect a small number of
prior demonstrations via teleoperation with a passive arm Gello \cite{gello}, that we modified for UR3e
controller, and then run different RL methods for a fixed
number of interaction steps. All methods use the exact same
hyper-parameters and network architectures as in metaworld
tasks. We illustrate these tasks in Fig. \ref{real_world_result} and briefly describe
them here.

\textbf{Lift:} The objective is to pick up a foam block. The initial
location of the block is randomized over roughly \( 22cm \times (20cm - 25cm)\) 
trapezoid, which covers the entire area visible
from the wrist-camera when the robot is at the home position. See Fig.\ref{lift}, Fig. \ref{lift_rand} for reference.
We collect 10 demonstrations for this task due to its simplicity.
It uses wrist-camera images as observations. We detect whether the gripper is holding the block by checking if the gripper
width is static and the desired gripper width is smaller than the actual gripper width. The success detector returns 1 if the end
effector has moved upward by at least 2cm while holding the block. The maximum episode length is 120.

\textbf{Pick-and-Place}: The objective is to pick a soft toy on
a wooden block. The initial location of the soft toy and the block
is in a fixed position. The soft toy is initialized so that it is visible from the wrist camera view, see Fig.\ref{2s_env}. We use 20 prior
demonstrations. This task uses environment camera images as
observations as well because the wrist-camera loses sight of the block
after picking up the soft toy. We detect whether the gripper is holding the soft toy by checking if the gripper
width is static and the desired gripper width is smaller than the actual gripper width. The success returns 1 when the gripper width is static, the end-effector position is near the threshold of desired block position. The maximum episode length is 300.

As the primary goal of our real-world evaluations is to
compare sample efficiency and performance of various algorithms, we design rule-based success detectors and perform
manual reset between episodes to ensure accurate reward and
initial conditions. Note that the sparse 0/1 reward
from the success detector is the only source of reward.

% \FloatBarrier % Ensures that the figure is placed before this barrier

\subsection{Real-world Experiment Results}

\textbf{Does \NAVINACT{} Work Effectively in Real-World Scenarios?}
Fig. \ref{real_world_result} presents the training curves of \NAVINACT{} compared to IBRL and across all different tasks. Each task is allowed different interactions based on its difficulty, and the success rate is measured over each 1000-step interval during training. \NAVINACT{} consistently learns faster than IBRL, and in the challenging task, Pick-and-Place, it is the only method that surpasses behavior cloning (BC) under 5K interactions.

In real-world evaluations, we assessed the final performance of each method using identical initial conditions. In the Lift task, \NAVINACT{} achieves the highest success rate in both regular and ``Hard Eval" settings, where the block's initial position is more challenging. Even when the task difficulty increases, \NAVINACT{} maintains a high success of 90\% rate, demonstrating its robustness to distributional shifts and unseen scenarios, unlike BC, which fails in harder settings due to its reliance on the static training dataset as shown in Fig. \ref{real_world_result}.

\textbf{How does \NAVINACT{} perform for long-horizon tasks?}
The Pick-and-Place task, the most difficult due to two-stage task manipulation, further highlights \NAVINACT{}'s advantages as displayed in Fig. \ref{real_world_result}. While BC also doesn't perform well due to suboptimal demonstration, other methods IBRL fail to match its success rate. While BC achieves a limited success rate of 20\%, \NAVINACT{} surpasses this by reaching a 75\% success rate, even in complex scenarios where small errors can lead to significant failures.

%% file: root/sections/6.conclusion.tex
\section{Discussion}
\label{section:conclusion}
\subsection{Conclusion}
In this paper, we presented \NAVINACT{}, a hierarchical policy framework that combines NavNet for waypoint motion-planning, ModeNet for dynamic mode switching, and InteractNet for fine-grained manipulation. By integrating imitation learning (IL) with reinforcement learning (RL), \NAVINACT{} effectively addresses challenges of sample efficiency and distributional shift, outperforming existing methods. Through extensive simulations and real-world experiments, we demonstrated \NAVINACT{}'s superior performance, generalizability, and adaptability across various tasks. Our approach learns faster and requires fewer interaction steps due to the integration of mode and waypoint prediction, making it a scalable and robust solution for complex robotic manipulation in real-world scenarios.

\subsection{Limitation and Future Work}
While \NAVINACT{} demonstrates significant advancements in sample efficiency and generalizability, one limitation of our approach is the reliance on collecting high-quality demonstration data, which can be time-consuming and expensive. Future work could focus on developing methods that allow non-experts to collect usable data quickly and efficiently, potentially through resources such as videos in the wild or collecting demonstrations using sketches.

%% file: main.bbl
\begin{thebibliography}{6}
%
% \bibitem{ros}
% S. Ross, G. Gordon, and D. Bagnell.: A reduction of imitation learning and structured prediction to no-regret online learning. In Proceedings of the fourteenth international conference
% on artificial intelligence and statistics, pages 627–635. JMLR Workshop and Conference Proceedings, 2011.
% \bibitem {smit:wat}
% Smith, T.F., Waterman, M.S.: Identification of common molecular subsequences.
% J. Mol. Biol. 147, 195?197 (1981). \url{doi:10.1016/0022-2836(81)90087-5}
% \bibitem{k_gandhi}
% Gandhi, Kanishk, Siddharth Karamcheti, Madeline Liao, and Dorsa Sadigh.: "Eliciting compatible demonstrations for multi-human imitation learning." In Conference on Robot Learning, pp. 1981-1991. PMLR, 2023.

% \bibitem{y_cui}
% Cui, Yuchen, David Isele, Scott Niekum, and Kikuo Fujimura.: "Uncertainty-aware data aggregation for deep imitation learning." In 2019 International Conference on Robotics and Automation (ICRA), pp. 761-767. IEEE, 2019.

% \bibitem{r_hoque}
% Hoque, Ryan, Ashwin Balakrishna, Ellen Novoseller, Albert Wilcox, Daniel S. Brown, and Ken Goldberg.: "Thriftydagger: Budget-aware novelty and risk gating for interactive imitation learning." arXiv preprint arXiv:2109.08273 (2021).

% \bibitem{m_kelly}
% Kelly, Michael, Chelsea Sidrane, Katherine Driggs-Campbell, and Mykel J. Kochenderfer.: "Hg-dagger: Interactive imitation learning with human experts." In 2019 International Conference on Robotics and Automation (ICRA), pp. 8077-8083. IEEE, 2019.

% \bibitem{m_mandlekar}
% Mandlekar, Ajay, Danfei Xu, Roberto Martín-Martín, Yuke Zhu, Li Fei-Fei, and Silvio Savarese.: "Human-in-the-loop imitation learning using remote teleoperation." arXiv preprint arXiv:2012.06733 (2020).

% \bibitem{m_l_schrum}
% Schrum, Mariah L., Erin Hedlund-Botti, Nina Moorman, and Matthew C. Gombolay.: "Mind meld: Personalized meta-learning for robot-centric imitation learning." In 2022 17th ACM/IEEE International Conference on Human-Robot Interaction (HRI), pp. 157-165. IEEE, 2022.

% \bibitem{laskey_et_al}
% Laskey, Michael, Jonathan Lee, Roy Fox, Anca Dragan, and Ken Goldberg.: "Dart: Noise injection for robust imitation learning." In Conference on robot learning, pp. 143-156. PMLR, 2017.

\bibitem{zhu}
Zhu, Yifeng, Abhishek Joshi, Peter Stone, and Yuke Zhu.: ``Viola: Imitation learning for vision-based manipulation with object proposal priors." In Conference on Robot Learning, pp. 1199-1210. PMLR, 2023.

\bibitem{s_karamcheti}
Karamcheti, Siddharth, Suraj Nair, Annie S. Chen, Thomas Kollar, Chelsea Finn, Dorsa Sadigh, and Percy Liang.: ``Language-driven representation learning for robotics." arXiv preprint arXiv:2302.12766 (2023).

\bibitem{s_nair}
Nair, Suraj, Aravind Rajeswaran, Vikash Kumar, Chelsea Finn, and Abhinav Gupta.: ``R3m: A universal visual representation for robot manipulation." arXiv preprint arXiv:2203.12601 (2022).

\bibitem{pertsch}
Pertsch, Karl, Youngwoon Lee, and Joseph Lim.: ``Accelerating reinforcement learning with learned skill priors." In Conference on robot learning, pp. 188-204. PMLR, 2021.


\bibitem{lava}
Bhaskar, Amisha, Rui Liu, Vishnu D. Sharma, Guangyao Shi, and Pratap Tokekar. : "Lava: Long-horizon visual action based food acquisition." arXiv preprint arXiv:2403.12876 (2024).

\bibitem{pegMARL}
Yu, Peihong, Manav Mishra, Alec Koppel, Carl Busart, Priya Narayan, Dinesh Manocha, Amrit Bedi, and Pratap Tokekar. : "Beyond Joint Demonstrations: Personalized Expert Guidance for Efficient Multi-Agent Reinforcement Learning." arXiv preprint arXiv:2403.08936 (2024).
% \bibitem{avil}
% Liu, Rui, Amisha Bhaskar, and Pratap Tokekar. "Adaptive visual imitation learning for robotic assisted feeding across varied bowl configurations and food types." arXiv preprint arXiv:2403.12891 (2024).



\bibitem{lynch}
Lynch, Corey, Mohi Khansari, Ted Xiao, Vikash Kumar, Jonathan Tompson, Sergey Levine, and Pierre Sermanet.: ``Learning latent plans from play." In Conference on robot learning, pp. 1113-1132. PMLR, 2020.



\bibitem{belkhale}
Belkhale, Suneel, and Dorsa Sadigh.: ``Plato: Predicting latent affordances through object-centric play." In Conference on Robot Learning, pp. 1424-1434. PMLR, 2023.


\bibitem{shridhar}
Shridhar, Mohit, Lucas Manuelli, and Dieter Fox.: ``Perceiver-actor: A multi-task transformer for robotic manipulation." In Conference on Robot Learning, pp. 785-799. PMLR, 2023.
\bibitem{rlpd}
Ball, Philip J., Laura Smith, Ilya Kostrikov, and Sergey Levine.: ``Efficient online reinforcement learning with offline data." In International Conference on Machine Learning, pp. 1577-1594. PMLR, 2023.

\bibitem{akgun}
Akgun, Baris, Maya Cakmak, Karl Jiang, and Andrea L. Thomaz.: ``Keyframe-based learning from demonstration: Method and evaluation." International Journal of Social Robotics 4 (2012): 343-355.


\bibitem{chi}
Chi, Cheng, Siyuan Feng, Yilun Du, Zhenjia Xu, Eric Cousineau, Benjamin Burchfiel, and Shuran Song.: ``Diffusion policy: Visuomotor policy learning via action diffusion." arXiv preprint arXiv:2303.04137 (2023).


\bibitem{lee_et_al}
Lee, Michelle A., Carlos Florensa, Jonathan Tremblay, Nathan Ratliff, Animesh Garg, Fabio Ramos, and Dieter Fox.: ``Guided uncertainty-aware policy optimization: Combining learning and model-based strategies for sample-efficient policy learning." In 2020 IEEE international conference on robotics and automation (ICRA), pp. 7505-7512. IEEE, 2020.

\bibitem{yamada}
Yamada, Jun, Youngwoon Lee, Gautam Salhotra, Karl Pertsch, Max Pflueger, Gaurav Sukhatme, Joseph Lim, and Peter Englert.: ``Motion planner augmented reinforcement learning for robot manipulation in obstructed environments." In Conference on Robot Learning, pp. 589-603. PMLR, 2021.

\bibitem{cheng}
Cheng, Shuo, and Danfei Xu.: ``League: Guided skill learning and abstraction for long-horizon manipulation." IEEE Robotics and Automation Letters (2023).

\bibitem{xia_et_at}
Xia, Fei, Chengshu Li, Roberto Martín-Martín, Or Litany, Alexander Toshev, and Silvio Savarese.: ``Relmogen: Leveraging motion generation in reinforcement learning for mobile manipulation." arXiv preprint arXiv:2008.07792 (2020).

\bibitem{james_davison}
James, Stephen, and Andrew J. Davison.: ``Q-attention: Enabling efficient learning for vision-based robotic manipulation." IEEE Robotics and Automation Letters 7, no. 2 (2022): 1612-1619.

\bibitem{james_et_al}
James, Stephen, Kentaro Wada, Tristan Laidlow, and Andrew J. Davison.: ``Coarse-to-fine q-attention: Efficient learning for visual robotic manipulation via discretisation." In Proceedings of the IEEE/CVF Conference on Computer Vision and Pattern Recognition, pp. 13739-13748. 2022.

\bibitem{liu_et_al}
Liu, I-Chun Arthur, Shagun Uppal, Gaurav S. Sukhatme, Joseph J. Lim, Peter Englert, and Youngwoon Lee.: ``Distilling motion planner augmented policies into visual control policies for robot manipulation." In Conference on Robot Learning, pp. 641-650. PMLR, 2022.


\bibitem{red_q}
Chen, Xinyue, Che Wang, Zijian Zhou, and Keith Ross.: ``Randomized ensembled double q-learning: Learning fast without a model." arXiv preprint arXiv:2101.05982 (2021).


\bibitem{silver}
Silver, David, Julian Schrittwieser, Karen Simonyan, Ioannis Antonoglou, Aja Huang, Arthur Guez, Thomas Hubert et al.,: ``Mastering the game of go without human knowledge." nature 550, no. 7676 (2017): 354-359.

\bibitem{oriol}
Vinyals, Oriol, Igor Babuschkin, Wojciech M. Czarnecki, Michaël Mathieu, Andrew Dudzik, Junyoung Chung, David H. Choi et al.,: ``Grandmaster level in StarCraft II using multi-agent reinforcement learning." nature 575, no. 7782 (2019): 350-354.

\bibitem{fair}
Meta Fundamental AI Research Diplomacy Team (FAIR)†, Anton Bakhtin, Noam Brown, Emily Dinan, Gabriele Farina, Colin Flaherty, Daniel Fried et al.,: ``Human-level play in the game of Diplomacy by combining language models with strategic reasoning." Science 378, no. 6624 (2022): 1067-1074.

\bibitem{dropout_q}
Hiraoka, Takuya, Takahisa Imagawa, Taisei Hashimoto, Takashi Onishi, and Yoshimasa Tsuruoka.: ``Dropout q-functions for doubly efficient reinforcement learning." arXiv preprint arXiv:2110.02034 (2021).

\bibitem{denis}
Yarats, Denis, Rob Fergus, Alessandro Lazaric, and Lerrel Pinto.: ``Mastering visual continuous control: Improved data-augmented reinforcement learning." arXiv preprint arXiv:2107.09645 (2021).

\bibitem{Kumar}
Kumar, Aviral, Anikait Singh, Frederik Ebert, Mitsuhiko Nakamoto, Yanlai Yang, Chelsea Finn, and Sergey Levine. : ``Pre-training for robots: Offline rl enables learning new tasks from a handful of trials." arXiv preprint arXiv:2210.05178 (2022).

\bibitem{Brohan} 
Brohan, Anthony, Noah Brown, Justice Carbajal, Yevgen Chebotar, Joseph Dabis, Chelsea Finn, Keerthana Gopalakrishnan et al., :``Rt-1: Robotics transformer for real-world control at scale." arXiv preprint arXiv:2212.06817 (2022).


\bibitem{ibrl} 
Hu, Hengyuan, Suvir Mirchandani, and Dorsa Sadigh.: ``Imitation Bootstrapped Reinforcement Learning." arXiv preprint arXiv:2311.02198 (2023).

\bibitem{hydra} 
Belkhale, Suneel, Yuchen Cui, and Dorsa Sadigh. ``Hydra: Hybrid robot actions for imitation learning." In Conference on Robot Learning, pp. 2113-2133. PMLR, 2023.

\bibitem{planseq} 
Dalal, Murtaza, Tarun Chiruvolu, Devendra Chaplot, and Ruslan Salakhutdinov. ``Plan-seq-learn: Language model guided rl for solving long horizon robotics tasks." arXiv preprint arXiv:2405.01534 (2024).

\bibitem{rlpd}
Ball, Philip J., Laura Smith, Ilya Kostrikov, and Sergey Levine. ``Efficient online reinforcement learning with offline data." In International Conference on Machine Learning, pp. 1577-1594. PMLR, 2023.


\bibitem{hester}
Hester, Todd, Matej Vecerik, Olivier Pietquin, Marc Lanctot, Tom Schaul, Bilal Piot, Dan Horgan et al., ``Deep q-learning from demonstrations." In Proceedings of the AAAI conference on artificial intelligence, vol. 32, no. 1. 2018.

\bibitem{nair_ashvin}
Nair, Ashvin, Bob McGrew, Marcin Andrychowicz, Wojciech Zaremba, and Pieter Abbeel. ``Overcoming exploration in reinforcement learning with demonstrations." In 2018 IEEE international conference on robotics and automation (ICRA), pp. 6292-6299. IEEE, 2018.

\bibitem{complex_dexterous}
Rajeswaran, Aravind, Vikash Kumar, Abhishek Gupta, Giulia Vezzani, John Schulman, Emanuel Todorov, and Sergey Levine. ``Learning complex dexterous manipulation with deep reinforcement learning and demonstrations." arXiv preprint arXiv:1709.10087 (2017).


\bibitem{task_level}
Seo, Younggyo, Danijar Hafner, Hao Liu, Fangchen Liu, Stephen James, Kimin Lee, and Pieter Abbeel. ``Masked world models for visual control." In Conference on Robot Learning, pp. 1332-1344. PMLR, 2023.


\bibitem{gello}
Wu, Philipp, Yide Shentu, Zhongke Yi, Xingyu Lin, and Pieter Abbeel. ``Gello: A general, low-cost, and intuitive teleoperation framework for robot manipulators." arXiv preprint arXiv:2309.13037 (2023).


\bibitem{metaworld}
Yu, Tianhe, Deirdre Quillen, Zhanpeng He, Ryan Julian, Karol Hausman, Chelsea Finn, and Sergey Levine. ``Meta-world: A benchmark and evaluation for multi-task and meta reinforcement learning." In Conference on robot learning, pp. 1094-1100. PMLR, 2020.

\bibitem{td3}
Dankwa, Stephen, and Wenfeng Zheng. ``Twin-delayed ddpg: A deep reinforcement learning technique to model a continuous movement of an intelligent robot agent." In Proceedings of the 3rd international conference on vision, image and signal processing, pp. 1-5. 2019.

\bibitem{AT}
Strub, Marlin P., and Jonathan D. Gammell.: ``Adaptively Informed Trees (AIT*): Fast asymptotically optimal path planning through adaptive heuristics." In 2020 IEEE International Conference on Robotics and Automation (ICRA), pp. 3191-3198. IEEE, 2020.

\bibitem{hierarchial RL}
Hutsebaut-Buysse, Matthias, Kevin Mets, and Steven Latré. ``Hierarchical reinforcement learning: A survey and open research challenges." Machine Learning and Knowledge Extraction 4, no. 1 (2022): 172-221.

\end{thebibliography}
